\documentclass{article}

\PassOptionsToPackage{numbers, compress}{natbib}
\usepackage[main, final]{neurips_2026}
\usepackage[utf8]{inputenc} % allow utf-8 input
\usepackage[T1]{fontenc}    % use 8-bit T1 fonts
\usepackage{hyperref}       % hyperlinks
\usepackage{url}            % simple URL typesetting
\usepackage{booktabs}       % professional-quality tables
\usepackage{amsfonts}       % blackboard math symbols
\usepackage{nicefrac}       % compact symbols for 1/2, etc.
\usepackage{microtype}      % microtypography
\usepackage{xcolor}         % colors

\usepackage{amssymb} 

\title{HilbertGen-3D:Hilbert--Multifractal Conditioning for Topology-Aware 3D Generation}

\usepackage{etoolbox, amsmath,amsfonts,amsbsy,amsgen,amscd,mathrsfs,amssymb,amsthm,bm, mathtools}
\usepackage{titlesec}
\usepackage{dsfont}
\usepackage{algorithm}
\usepackage[noend]{algorithmic}
\usepackage{enumitem}
\usepackage{caption,wrapfig}
\usepackage{subcaption}
\hypersetup{citecolor=[RGB]{0,0,255}}
\usepackage{booktabs} 
\usepackage{cite}
\usepackage{footnotehyper}
\makesavenoteenv{table}
\makesavenoteenv{tabular}

\usepackage{algorithm}
\usepackage{algorithmic}

\usepackage{cleveref}
\allowdisplaybreaks

\let\epsilon\varepsilon

\newboolean{showcomment}
\setboolean{showcomment}{true}
\newcommand{\todo}[1]{\ifthenelse{\boolean{showcomment}}{{\textcolor{red}{\bf Todo:  #1}}}{}}
\newcommand{\guannan}[1]{\ifthenelse{\boolean{showcomment}}{{\textcolor{red}{\bf [Guannan:  #1]}}}{}}

\usepackage{tikz}
\usetikzlibrary{matrix,decorations.pathreplacing,calc}

\usepackage{multirow}
\usepackage{algorithm,algorithmic}

\pgfkeys{tikz/mymatrixenv/.style={decoration=brace,every left delimiter/.style={xshift=3pt},every right delimiter/.style={xshift=-3pt}}}
\pgfkeys{tikz/mymatrix/.style={matrix of math nodes,left delimiter=[,right delimiter={]},inner sep=2pt,column sep=1em,row sep=0.5em,nodes={inner sep=0pt}}}
\pgfkeys{tikz/mymatrixbrace/.style={decorate,thick}}

\usepackage{graphicx} % for \resizebox
\usepackage[table]{xcolor}

\author{%
  Ziyi Zhang$^{1,*}$, Shixuan Li$^{2,*}$,\\
  \textbf{Yorie Nakahira$^{1}$, Paul Bogdan$^{2}$, Guannan Qu$^{1}$}\\
  \small
  $^{1}$Department of Electrical and Computer Engineering, Carnegie Mellon University\\
  \small
  $^{2}$Ming Hsieh Department of Electrical and Computer Engineering, University of Southern California\\
}

\begin{document}

\maketitle
\def\thefootnote{*}\footnotetext{These authors contributed equally to this work.}

\begin{abstract}
  High-fidelity 3D generative modeling underpins game asset creation, medical imaging, and VR/AR environmental design, where geometric structure directly influences perception and decision-making. Diffusion models represent a powerful paradigm for point cloud generation. However, existing approaches rely heavily on local voxel or patch features and often miss global structural organization, and much less work has studied how global conditioning can be used to improve generation quality. In this paper, we introduce HilbertGen-3D, a novel topology-aware conditional diffusion framework based on Hilbert–Multifractal Conditioning (HMC). HMC computes multi-scale voxel measures, serializes them via a 3D Hilbert curve to expose structured spatial dependencies, and estimates the generalized fractal dimensions from scale-dependent partition functions. We encode these signals as global embeddings and multi-scale tokens, and inject them into a diffusion denoiser through an efficient bottleneck fusion module. Across the Chair, Airplane, and Car categories of ShapeNet, our two conditioning variants reduce the absolute deviation of 1-Nearest Neighbor Accuracy (1-NNA) from 50\% by up to 8.90 percentage points and increase coverage by up to 5.88 percentage points relative to TopoDiT-3D~\citep{Guan25}. This shows that combining local density cues with global Hilbert–multifractal structure provides a simple and effective recipe for high-fidelity and diverse 3D generation. Code is available at \url{https://github.com/shixuanli01/HMC-DiT3D}.
\end{abstract}

\section{Introduction}
%Accurate and efficient three-dimensional (3D) representations of the world are fundamental to enabling autonomous systems to perceive, plan, and act in dynamic, unstructured, and uncertain environments, bridging the gap between raw sensor data and the geometric understanding required for robust control and robotic manipulation. 
Generative modeling of three-dimensional (3D) point clouds is a central problem in geometric learning, with applications spanning shape synthesis, digital content creation, simulation, and 3D perception~\citep{Hao_2021_ICCV, Litany2017DeformableSC}. Diffusion models have recently become powerful tools for this task because they can represent highly multimodal shape distributions and scale to expressive neural backbones~\citep{Gao2023MaskedDT}. Nevertheless, many existing architectures, such as DiT3D~\citep{mo2023dit3d}, are still primarily driven by local geometric operators: they voxelize or patchify the shape, construct local tokens, and rely on the denoiser to recover global point distribution. In practice, this can produce shapes with locally plausible geometry but weaker global coherence, especially for objects with thin parts, holes, branches, or strongly nonuniform mass distributions across scales.

This limitation suggests that global structural information should explicitly guide the denoising process, rather than being left to emerge from local feature aggregation.
%One approach to realize this idea is TopoDiT-3D~\citep{Guan25}, which augments a diffusion transformer with persistence images derived from persistent homology and a bottleneck fusion module. %However, persistent homology conditioning requires a complex pipeline for extracting and encoding topological summaries. This motivates novel conditioning mechanisms informing global structures that are easy to compute and integrate into standard diffusion backbones.
In this paper, we propose \emph{HilbertGen-3D}, a novel topology-aware framework for 3D point cloud generation based on Hilbert–Multifractal Conditioning (HMC). Instead of constructing persistence diagrams, we start from multi-scale voxel measures computed on a normalized point cloud. At each scale, we serialize the voxel grid using a 3D Hilbert mapping, converting the spatial mass distribution into a structured one-dimensional sequence. We then %apply 
exploit the multifractal %methodology of 
theory \citep{mandelbrot2013multifractals,Salat17,zhang2025samplingcomplexityawareframeworkdiscretetime,zhang2026learningfractionalorderdynamicssingle} and evaluate the partition function across scales and moments to estimate the generalized fractal dimensions summarizing how mass concentrates, disperses, and varies across resolutions. Together with the Hilbert-ordered sequences themselves, these statistics define an HMC representation of the shape.

HMC provides two complementary forms of conditioning. A compact global descriptor captures the overall multifractal profile of the shape, while windowed embeddings of the Hilbert sequences retain structured multi-scale context. We encode these signals into a global embedding and a set of HMC conditioning tokens, then inject them into a diffusion denoiser via a memory-efficient bottleneck fusion module. This design uses a representation built directly from voxel mass statistics to preserve the key insight behind topology-aware conditioning, allowing for global shape information to guide denoising. We also verify that HMC improves structural fidelity, sample diversity, and global coherence relative to purely local baselines on standard point-cloud generation benchmarks.

\section{Related Works}
\paragraph{3D generative models.}
Deep generative modeling for 3D shapes has advanced across a range of representations,
including point clouds, voxels, meshes, and implicit fields~\citep{xu2023survey}. Early
point-set methods, such as PointFlow~\citep{yang2019pointflow}, modeled shape distributions with hierarchical
continuous normalizing flows, while ShapeGF~\citep{cai2020shapegf} learned gradient fields whose sampling
procedure recovers surface geometry. Diffusion
models then became a dominant paradigm for 3D shape synthesis: PVD~\citep{Zhou20213DSG} formulated
denoising diffusion directly on point clouds, LION~\citep{zeng2022lion} moved diffusion to a hierarchical latent
space to improve fidelity and enable downstream manipulations, and DiT-3D~\citep{mo2023dit3d} showed that
plain diffusion transformers can serve as competitive 3D generators. In parallel, conditional systems such
as Point-E~\citep{nichol2022pointe} and Shap-E~\citep{jun2023shape} expanded generation beyond unconditional shape synthesis, while
GET3D~\citep{gao2022get3d} demonstrated high-quality textured mesh generation from image collections. More recently, topology-aware 3D diffusion has begun to incorporate persistent homology(PH) features as global structural cues~\citep{Guan25}. 
%Our contribution is not to reject this insight, but to replace persistent homology-derived topology tokens with a conditioning signal that is more directly tied to point cloud reconstruction. %We introduce Hilbert--Multifractal Conditioning (HMC),
This paper introduces a new approach based on the Hilbert–Multifractal Conditioning method, which injects Hilbert-serialized multifractal descriptors as multi-scale
structural priors for point-cloud generation. Whereas persistent homology captures connected components, loops, and voids, HMC captures voxel mass distribution, density heterogeneity, and local spatial neighborhoods. Furthermore, persistent homology uses a dedicated topological encoding pipeline, while Hilbert sequences can be directly processed by sequence models and bottleneck fusion modules, which can exploit prior information about the reference density structure to improve generation quality and diversity.

\paragraph{Hilbert curves and structure-preserving serialization for 3D data.} Space-filling curves provide a principled way to analyze multidimensional geometry. % while approximately preserving locality. 
Classical analysis showed that the Hilbert curve has
strong clustering behavior and can preserve locality better than the z-curve in common query
settings~\citep{moon2001hilbert}. This idea has recently reappeared in 3D deep learning as a
practical serialization tool for irregular point clouds. OctFormer~\citep{wang2023octformer} uses octree-based ordering
to form efficient local windows for transformer processing.
Point Transformer V3 explicitly serializes point clouds with z-order and Hilbert curves to
build efficient grouped attention, arguing that Hilbert ordering offers stronger locality
preservation~\citep{wu2024ptv3}. PointMamba~\citep{liang2024pointmamba} similarly leverages space-filling curves to
turn unstructured point clouds into sequences amenable to linear-time state-space modeling. Point2Point~\citep{pandhare2023point2point} goes further by directly learning on
Hilbert-sorted point clouds and reports competitive results on both analysis and generation
tasks. These works use Hilbert order primarily as a
computational device for tokenization, grouping, or sequence modeling. In contrast, our
method uses the Hilbert traversal as part of the conditioning signal itself: we serialize
multi-scale voxel measures along the curve, extract both global multifractal summaries and
local Hilbert-window embeddings, and feed these descriptors into the denoiser as explicit
global structural priors.

\section{Preliminaries}
%We first review denoising diffusion probabilistic models for 3D point cloud generation, and then introduce the Hilbert curve as the locality-preserving space-filling curve used in our conditioning framework.

We review denoising diffusion probabilistic models for 3D point cloud generation and then introduce the Hilbert curve as the locality-preserving space-filling curve used in our conditioning framework.

\subsection{Denoising Diffusion Probabilistic Model (DDPM) for 3D Point Clouds}
\label{sec:DDPM}

Following ~\citet{mo2023dit3d} and ~\citet{Guan25},
we model 3D point cloud generation using a conditional denoising diffusion
probabilistic model (DDPM)~\citep{Ho20denoise}. Let
$X_0\in\mathbb R^{N\times3}$ denote a clean point cloud containing $N$
points, and let $\mathcal C$ denote its associated conditioning information.
In our model, $\mathcal C$ includes the Hilbert--Multifractal Conditioning
(HMC) representation introduced in Section~4. Gaussian distributions and
Euclidean norms below are understood on the $3N$ vectorized point
coordinates, with $I$ denoting the corresponding identity matrix.

The forward process gradually perturbs $X_0$ using a prescribed variance
schedule $\{\beta_t\}_{t=1}^T\subset(0,1)$:
\[
q(X_{1:T}\mid X_0)
=\prod_{t=1}^T q(X_t\mid X_{t-1}),
\qquad
q(X_t\mid X_{t-1})
=\mathcal N\!\left(X_t;\sqrt{\alpha_t}X_{t-1},\beta_t I\right),
\]
where $\alpha_t=1-\beta_t$. Writing
$\bar\alpha_t=\prod_{i=1}^t\alpha_i$ and $\bar\alpha_0=1$, we can sample
any intermediate state directly as
\[
X_t=\sqrt{\bar\alpha_t}X_0+\sqrt{1-\bar\alpha_t}\,\epsilon,
\qquad \epsilon\sim\mathcal N(0,I).
\]
The schedule is chosen so that the terminal distribution is approximately
standard Gaussian.

Generation reverses this corruption through a learned conditional Markov
chain:
\[
\begin{aligned}
p_\theta(X_{0:T}\mid\mathcal C)
&=p(X_T)\prod_{t=1}^T p_\theta(X_{t-1}\mid X_t,\mathcal C),\\
p_\theta(X_{t-1}\mid X_t,\mathcal C)
&=\mathcal N\!\left(
X_{t-1};\mu_\theta(X_t,t,\mathcal C),\sigma_t^2 I
\right),
\end{aligned}
\]
where $p(X_T)=\mathcal N(0,I)$ and $\sigma_t^2$ is a prescribed reverse
variance. For $t\geq2$, common choices are $\sigma_t^2=\beta_t$ and
$\sigma_t^2=\beta_t(1-\bar\alpha_{t-1})/(1-\bar\alpha_t)$~\citep{Ho20denoise}.
Rather than predicting the reverse mean directly, we use a neural network
$\epsilon_\theta(X_t,t,\mathcal C)\in\mathbb R^{N\times3}$ to predict
the injected noise, and parameterize the mean as~\citep{Ho20denoise}
\[
\mu_\theta(X_t,t,\mathcal C)
=\frac{1}{\sqrt{\alpha_t}}
\left(
X_t-\frac{\beta_t}{\sqrt{1-\bar\alpha_t}}
\epsilon_\theta(X_t,t,\mathcal C)
\right).
\]

The conditional log-likelihood admits the variational lower
bound~\citep{Ho20denoise}:
\[
\log p_\theta(X_0\mid\mathcal C)
\geq
\mathbb E_{q(X_{1:T}\mid X_0)}
\left[
\log\frac{p_\theta(X_{0:T}\mid\mathcal C)}
{q(X_{1:T}\mid X_0)}
\right].
\]
With fixed reverse variances, the denoising KL terms for $t\geq2$ in the
negative bound reduce to time-weighted squared errors in noise prediction.
Following the simplified DDPM objective~\citep{Ho20denoise}, we instead
use uniform weighting across timesteps:
\[
\mathcal L_{\mathrm{DDPM}}
=\mathbb E_{(X_0,\mathcal C),t,\epsilon}
\left[
\left\|\epsilon-\epsilon_\theta(X_t,t,\mathcal C)\right\|_2^2
\right],
\]
where $(X_0,\mathcal C)$ is a training shape--condition pair,
$t\sim\operatorname{Unif}\{1,\ldots,T\}$, and
$\epsilon\sim\mathcal N(0,I)$ is sampled independently.
Thus, the implemented noise-prediction loss is a reweighted surrogate
rather than the unchanged variational objective. For each $t$, the population minimizer of the squared-error objective
predicts $\mathbb E[\epsilon\mid X_t,\mathcal C]$.
The rescaled prediction
$-\epsilon_\theta(X_t,t,\mathcal C)/\sqrt{1-\bar\alpha_t}$
therefore estimates this score, providing a score-matching interpretation
of the same denoiser~\citep{Ho20denoise}.
After training, we sample $X_T\sim\mathcal N(0,I)$ and apply the learned
reverse transitions for $t=T,\ldots,1$, holding $\mathcal C$ fixed, to
obtain a generated point cloud $X_0$.

\subsection{Hilbert Curves}

\begin{figure}[ht!]
    \centering
    \includegraphics[width=\linewidth]{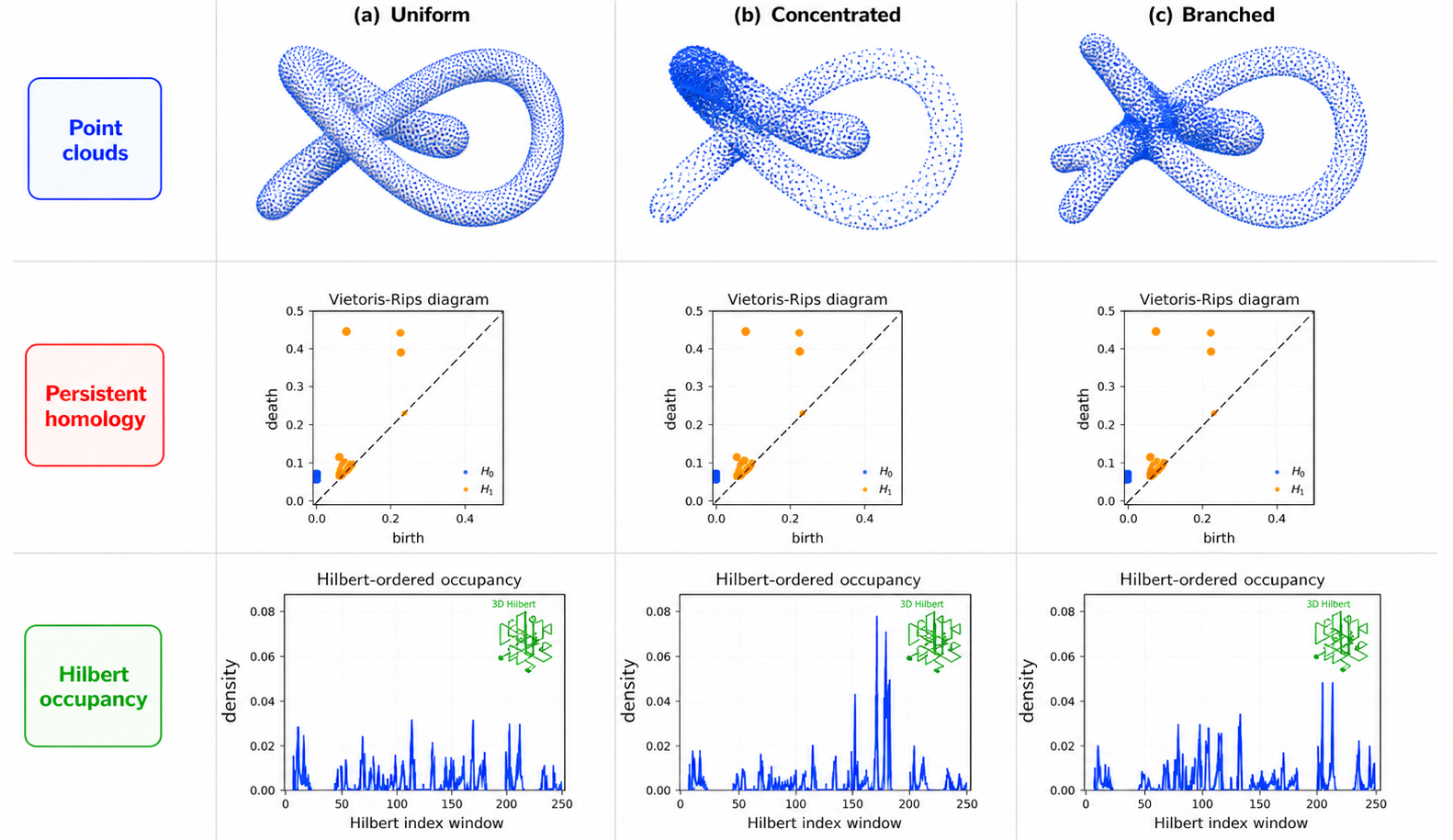}
    \caption{Comparison of point clouds (top row), their persistence diagrams (middle row), and Hilbert-ordered mass profiles associated with the sequences \(c_s[t]\). The examples illustrate changes in spatial mass concentration despite similar coarse topological structure.}
    \label{fig:Hilbert_illustration}
\end{figure}

Our denoiser uses the diffusion-transformer backbone of DiT-3D~\citep{mo2023dit3d}. To construct a separate sequence of geometric conditioning signals, we serialize a voxel mass field using a 3D Hilbert traversal. This orders voxel cells, rather than the original points, and provides a spatially structured input to the HMC condition encoder.
A Hilbert curve provides such an ordering. In the continuous setting, the Hilbert curve is a
space-filling map
\begin{equation*}
\gamma:[0,1]\rightarrow [0,1]^d.
\end{equation*}
For our purposes, we fix $d=3$ in the rest of the paper. Hilbert curves are constructed recursively: a unit cube is divided into
$2^d$ smaller subcubes, and at a finer level, each octant is recursively traversed using an appropriately rotated or reflected lower-order pattern. Its orientation is chosen so that the exit from one octant connects to the entry into the next; the same unrotated pattern is not simply repeated independently in every octant.. For more detail on the construction, see~\citet{moon2001hilbert}. As the recursion depth increases, the Hilbert curve visits finer and finer spatial cells, and in the limit, it fills the entire domain. For computation, we use the discrete 3D Hilbert traversal. At scale $s$, let
$R_s=2^s$ and $V_s=\{0,\ldots,R_s-1\}^3$ be a regular voxel grid. A discrete Hilbert ordering is a bijection
\begin{equation*}
\pi_s:V_s\rightarrow \{0,\ldots,R_s^3-1\},
\end{equation*}
which assigns each voxel a unique one-dimensional index and visits every voxel exactly once.
The inverse map $\pi_s^{-1}$ maps a Hilbert index back to its corresponding voxel cell. Thus, Hilbert serialization only permutes the voxel values and therefore introduces no additional information loss beyond voxelization.

\textbf{Locality preservation.} The key property of the Hilbert curve is locality preservation. Although no one-dimensional
ordering can perfectly preserve all Euclidean distances in three dimensions, Hilbert order tends to
place spatially nearby voxels close to each other in the induced sequence~\citep{moon2001hilbert}. Contiguous intervals of
Hilbert indices often correspond to compact spatial neighborhoods, and the recursive construction
makes this property hold across multiple resolutions. Compared with simple raster or lexicographic
ordering, Hilbert ordering avoids many long spatial jumps and better preserves clustered structure. 
This property is especially useful for point-cloud generation. After a point cloud is normalized and
voxelized, the Hilbert traversal converts its spatial mass distribution into an ordered signal. Coarse
recursion levels encode global layout, while deeper recursion levels resolve local details such as
thin parts, branches, and nonuniform density. Therefore, Hilbert serialization provides a natural
interface between unordered 3D geometry and sequence-based neural architectures. In our method,
this ordering is used to construct Hilbert--Multifractal Conditioning tokens that provide the denoiser
with locality-preserving, multi-scale structural information.

\section{Methodology}
\label{section:method}
%\guannan{We need a problem formulation section? }

\textbf{Overview of HilbertGen-3D.} 
We use point clouds
$X \in \mathbb{R}^{N \times 3}$ , where $N$ represents the number of points with $x,y,z$ coordinates, sampled uniformly from a given surface to represent 3D shapes. HilbertGen-3D initially voxelizes it as input $v_i \in \mathbb{R}^{V\times V\times V \times 3}$ where $V$ represents the grid resolution. Following DiT-3D~\citep{mo2023dit3d}, the voxel grid is then divided into non-overlapping patches of size $p \times p \times p$, resulting in a sequence of patch tokens $s_i \in \mathbb{R}^{L\times d}$ with length $L = (V/p)^3$ and embedding dimension $d$ that serve as the primary input to the DiT backbone, where the embedding is realized via a learned linear projection. In \Cref{subsec:Hilbert}, we introduce locality-preserving density signals along Hilbert curves at different scales for local information and generalized dimension for global information, which we process through a lightweight transformer encoder to generate the HMC tokens. These tokens pass through a bottleneck
structure with multiple transformer blocks (DiT block) that employ self-attention to update the features, which we discuss in detail in \Cref{subsec:diffusion_integration}. Finally, to achieve denoising in the point cloud space, the voxelized representation is converted back to the original point cloud format via devoxelization, yielding the denoised point cloud. 

\subsection{Hilbert serialization and generalized dimensions}
\label{subsec:Hilbert}
\paragraph{Multi-scale voxel measures.} Having introduced DDPM-based point-cloud generation and the basics of Hilbert curves, we now describe how multi-scale Hilbert occupancy sequences are converted into conditioning
signals for the diffusion denoiser. We first generate the probability measure from the Hilbert curves. Choose a set of scales $\mathcal{S}=\{s_1,\dots,s_K\}$ with grid resolution $R_s = 2^{s}, s\in\mathcal{S}$ and voxel side length $\varepsilon_s = 1/R_s$.
Let $V_s$ be the set of voxels at scale $s$.
Define a nonnegative mass per voxel $v\in V_s$:
\begin{equation}
\label{eqn:m}
m_s(v) = \sum_{n=1}^N \mathbf{1}\{x_n \in v\},
\end{equation}
where we compute the sample density in each voxel. We then normalize \eqref{eqn:m} to obtain a probability measure over voxels:
\begin{equation}
\mu_s(v) = \frac{m_s(v)}{\sum_{v'\in V_s} m_s(v')}.
\label{eq:mu_def}
\end{equation}

\begin{figure}
    \centering
    \includegraphics[width=\linewidth]{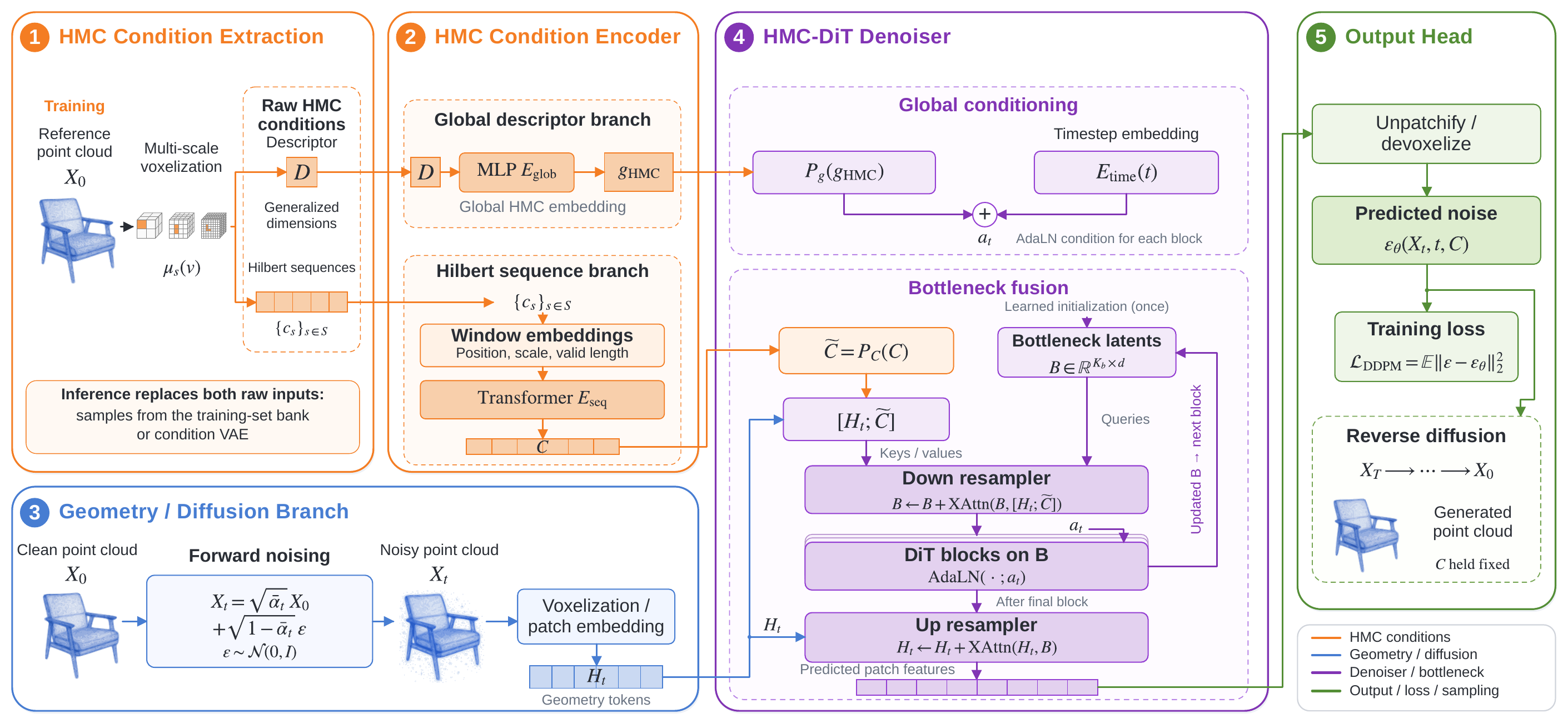}
    \caption{\textbf{Overview of HilbertGen-3D.}
    Multi-scale voxel measures are serialized along 3D Hilbert curves to construct Hilbert-ordered sequences, while generalized fractal dimensions form the global descriptor $D$. Separate encoders produce HMC tokens $C$ and a global HMC embedding $g_{\mathrm{HMC}}$, respectively. The noisy point cloud $X_t$ is embedded into geometry tokens $H_t$, which are fused with the projected HMC tokens through bottleneck latents $B$. Successive DiT blocks update $B$, with adaptive layer normalization conditioned on the timestep embedding $E_{\mathrm{time}}(t)$ and the projected global embedding $P_g(g_{\mathrm{HMC}})$. The processed latents update the geometry tokens, and the output head unpatchifies and devoxelizes the predicted features to obtain the noise estimate $\varepsilon_\theta(X_t,t,\mathcal{C})$ for DDPM training and reverse-diffusion sampling.}
    \label{fig:Structural}
\end{figure}

\paragraph{Hilbert serialization at each scale.}
Let $\pi_s:V_s \to \{0,\dots,|V_s|-1\}$ be a 3D Hilbert mapping,
and denote its inverse as $\pi_s^{-1}(t)$.
Define a Hilbert-ordered sequence:
\begin{equation}
c_s[t] \;=\; \mu_s\!\big(\pi_s^{-1}(t)\big), \qquad t=0,\dots,|V_s|-1.
\label{eq:hilbert_sequence}
\end{equation}

where we recover the point cloud density along the Hilbert curve. The bottom row of \Cref{fig:Hilbert_illustration} visualizes density variation along the Hilbert traversal associated with the sequence $c_s[t]$. Intuitively, $c_s[t]$ signals the point cloud's relative density and gives a rough indication of its overall shape. 

\paragraph{Partition functions and generalized dimensions.}
For a set of moments $\mathcal{Q}\subset\mathbb{R}$, define the scale-dependent partition function
\begin{equation}
Z_s(q) \;=\; \sum_{t=0}^{|V_s|-1} c_s[t]^q
\;=\; \sum_{v\in V_s} \mu_s(v)^q,
\qquad q\in\mathcal{Q}.
\label{eq:partition}
\end{equation}
Define the mass exponent $\tau(q)$ ~\citep{mandelbrot2013multifractals} that is obtained from the following formula: %via a log--log slope across scales:
\begin{equation}
\tau(q)=
\frac{\sum_{s\in\mathcal{S}}(\log\varepsilon_s-\overline{\log\varepsilon})(\log Z_s(q)-\overline{\log Z(q)})}
{\sum_{s\in\mathcal{S}}(\log\varepsilon_s-\overline{\log\varepsilon})^2},
\label{eq:tau_regression}
\end{equation}

where $\overline{\log\varepsilon} := \frac{1}{|\mathcal{S}|}\sum_s \log \epsilon_s$ and $\overline{\log Z(q)}) := \frac{1}{|\mathcal{S}|}\sum_s \log Z_s(q)$ denotes the sample average across scales in the least-squares.
Then the generalized fractal dimensions $D(q)$ are given by:
\begin{equation}
D(q) = \frac{\tau(q)}{q-1},\qquad q\neq 1.
\label{eq:Dq}
\end{equation}
Note that when $q = 1$, $D(q)$ can be computed via the entropy slope method; see e.g.~\citet{Salat17}, for details. The moment order $q$ controls which parts of the density distribution dominate the descriptor. Lower orders emphasize the spatial support of the shape: $q=1$ corresponds to an entropy-like information dimension, and $q=2$ is related to correlation dimension. Larger positive $q$ values emphasize high-density regions and therefore capture how density concentrates around thick parts or repeated structures. In point cloud generation, this gives the denoiser a compact summary of whether the object is uniformly distributed, concentrated into a few dense regions, or organized into thin, sparse structures. We further define an HMC descriptor vector by concatenating the generalized dimensions:
\begin{equation}
\mathbf D = \big(D(q)\big)_{q\in\mathcal{Q}} \in \mathbb{R}^{|\mathcal{Q}|}.
\label{eq:hmc_vector}
\end{equation}
We point out that the generalized dimensions summarize global scaling of the voxel density distribution and are invariant to voxel ordering. 

\paragraph{HMC condition encoding.}
The Hilbert-ordered sequences~\eqref{eq:hilbert_sequence} and HMC descriptors~\eqref{eq:hmc_vector} are encoded
through two separate branches. Each sequence $c_s$ is split into consecutive windows of length $w$,
with zero-padding of the final window when necessary. The window embeddings include position, scale, and valid-length information. A lightweight Transformer processes the concatenated multi-scale window
embeddings to produce the \textbf{HMC tokens}
$
C=E_{\mathrm{seq}}\!\left(\{c_s\}_{s\in\mathcal S}\right)
\in\mathbb R^{M\times d},$
where
$M=\sum_{s\in\mathcal S}
\left\lceil\frac{R_s^3}{w}\right\rceil$ and $E_{\mathrm{seq}}$ denotes the learned encoder.
The resulting embedding $C$ retains its window index, although self-attention can
incorporate context from other windows and scales. Separately, a
normalized multilayer perceptron embeds the descriptor into a \textbf{global HMC embedding}
$g_{\mathrm{HMC}}
=E_{\mathrm{glob}}(\mathbf D)\in\mathbb R^d.$
During training, both raw inputs are extracted from the clean shape.
At inference, a paired raw condition is sampled from the training-set
bank or generated by the condition VAE, then passed through the same
encoder. The complete conditioning input $\mathcal C$ in \Cref{sec:DDPM}
comprises both raw inputs.

\subsection{Integration into Diffusion Architectures}
\label{subsec:diffusion_integration}

We retain the voxel patchification and pointwise output structure of DiT-3D~\citep{mo2023dit3d} and use the bottleneck design of TopoDiT-3D~\citep{Guan25} to inject HMC conditions. As shown in \Cref{fig:Structural}, the denoiser receives geometry tokens \(H_t\in\mathbb R^{L\times d}\), HMC tokens \(C\in\mathbb R^{M\times d}\), and a global HMC embedding $g_{\mathrm{HMC}}\in\mathbb R^d$. The HMC tokens provide attention context via cross-attention, whereas the global embedding controls adaptive layer normalization. We first describe direct cross-attention as a reference formulation and then the bottleneck fusion used in our model.
\paragraph{Cross-attention conditioned on HMC tokens.}
In a Transformer denoiser block, let $H_t\in\mathbb{R}^{L\times d}$ be the current hidden tokens, and $C\in \mathbb{R}^{M \times d}$ the HMC tokens. As a simple conditioning baseline, HMC tokens can be injected into each denoising block by
cross-attention.
\begin{equation}
\mathrm{XAttn}(H_t,C)=\mathrm{softmax}\left(\frac{(H_t W_Q)(C W_K)^\top}{\sqrt{d}}\right)(C W_V),
\label{eq:xattn}
\end{equation}
and update
\begin{equation}
H_t \leftarrow H_t + \mathrm{XAttn}(H_t,C),
\label{eq:block_update}
\end{equation}

\textbf{Adaptive layer-norm conditioned on global HMC embedding.} Independently of the token-attention path, the HMC descriptor embedding modulates the denoiser through adaptive layer normalization. Let
\begin{equation}
\qquad a_t = E_{\mathrm{time}}(t) + P_g(g_{\mathrm{HMC}}),
\end{equation}
where $P_g$ is the learned affine projection, and $E_{\mathrm{time}}$ embeds the diffusion timestep.
For a token $h$, we use
\begin{equation}
\mathrm{AdaLN}(h; a_t)
=
\gamma(a_t) \odot \frac{h-\mu(h)}{\sigma(h)} + \beta(a_t),
\end{equation}
where $\mu(h)$ and $\sigma(h)$ are the mean and numerically stabilized
standard deviation across the token's features, and $\gamma(\cdot)$
and $\beta(\cdot)$ are learned scale and shift functions. 

The cross-attention
conditioning is easy to implement, but its cost scales with $LM$ in each conditioned block. It is therefore
most suitable when the number of HMC tokens is small. This motivates a bottleneck fusion design described below.

\paragraph{Bottleneck fusion via perceiver resampler}
Directly injecting all HMC tokens into the denoising transformer can be inefficient. The number of
Hilbert sequence tokens grow with the number of scales and with the voxel resolution, while voxelized point clouds are sparse and often produce many weakly informative patch tokens. Inspired
by the bottleneck design of TopoDiT-3D~\citep{Guan25}, we use a Perceiver-style resampler
to fuse local geometry tokens with global and multi-scale HMC tokens through a fixed-size latent
interface.

We introduce $K_b$ learnable
bottleneck latents $B \in \mathbb{R}^{K_b \times d}$, where $K_b \ll L+M$. The bottleneck first
compresses the concatenated local and HMC context:
\begin{equation}
B \leftarrow B + \mathrm{XAttn}\big(B,\;[H_t;\widetilde C]\big),
\label{eq:perceiver_down}
\end{equation}
where $\widetilde C=P_C(C)$ and $P_C$ is the learned projection. The resulting latent tokens $B$ are updated by DiT blocks using $a_t$ for AdaLN modulation. The processed $B$ is then used to update the geometry tokens:
\begin{equation}
H_t \leftarrow H_t + \mathrm{XAttn}\big(H_t,B\big).
\label{eq:perceiver_up}
\end{equation}
This module serves two purposes. First, it adaptively filters redundant sparse-voxel tokens and keeps the main DiT computation independent of the full HMC token length. Second, it provides a compact
fusion interface between local geometric features and Hilbert-multifractal structural priors. In this
way, the denoiser receives multi-scale shape information without attending directly to all local and condition tokens at every layer.

\section{Experiments}

\begin{figure}
    \centering
    \includegraphics[width=\linewidth]{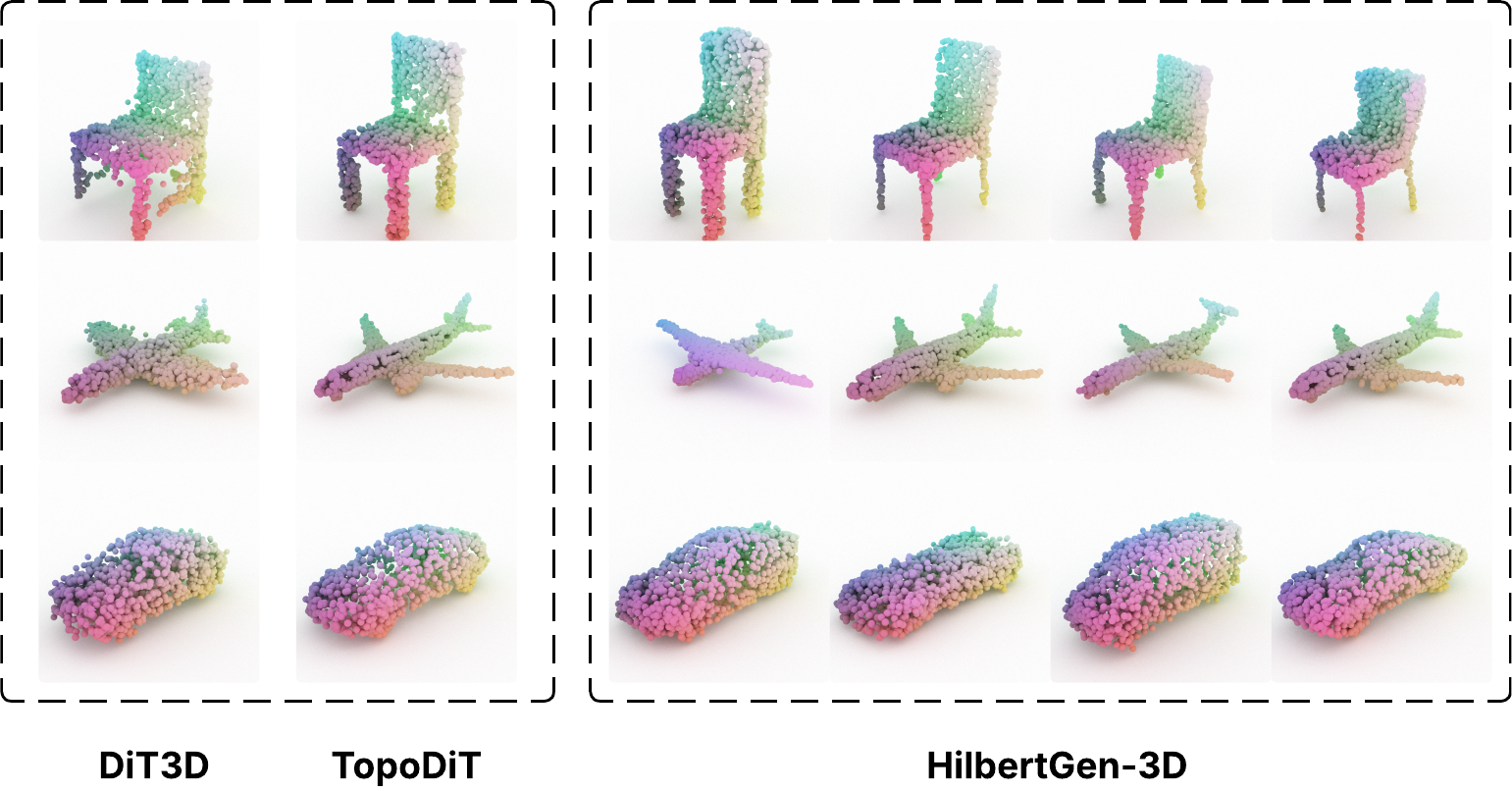}
    \caption{HilbertGen-3d results compared with DiT-3D and TopoDiT-3D. All experiments are conducted under same scale.}
    \label{fig:main-result}
\end{figure}

%In this section, we comprehensively evaluate HilbertGen-3D and conduct sensitivity experiments to investigate the effectiveness of the bottleneck structure and topological information within the point cloud data.

In this section, we evaluate HilbertGen-3D and conduct sensitivity experiments to investigate the effectiveness of the bottleneck structure and topological information within the point cloud data.

\subsection{Experiment Setup}
\paragraph{Dataset.} %In line with most of the previous works~\citep{yang2019pointflow,luo2021diffusion,achlioptas2018learning,Zhou20213DSG,zeng2022lion,mo2023dit3d,Guan25}, we use ShapeNet dataset as the benchmark for 3D point cloud generation~\citep{shapenet2015}. The ShapeNet benchmark comprises a dataset of $51,127$ shapes categorized into $55$ classes. Following prior work, we train HilbertGen-3D on the Chair, Airplane, and Car categories for 3D point cloud generation. We show further results on Lamp and Vessel in \Cref{appendix:additional_result}. 

In line with most prior work~\citep{yang2019pointflow,luo2021diffusion,achlioptas2018learning,Zhou20213DSG,zeng2022lion,mo2023dit3d,Guan25}, we use the ShapeNet dataset as the benchmark for 3D point cloud generation~\citep{shapenet2015}. ShapeNet comprises $51,127$ shapes across $55$ classes. Following prior work, we train HilbertGen-3D on the Chair, Airplane, and Car categories. We show further results on Lamp and Vessel in \Cref{appendix:additional_result}.

\paragraph{Evaluation.} %Following the prior works~\citep{mo2023dit3d,Guan25}, we use Chamfer Distance (CD) and Earth Movers Distance (EMD) as distance metrics to calculate 1-Nearest Neighbor Accuracy (1-NNA) and Coverage (COV) as evaluation metrics. 
Following prior work~\citep{mo2023dit3d,Guan25}, we use Chamfer Distance (CD) and Earth Mover’s Distance (EMD) to compute 1-Nearest Neighbor Accuracy (1-NNA) and Coverage (COV). The 1-NNA metric computes the leave-one-out accuracy of a 1-Nearest Neighbor classifier, with lower scores signifying superior generative performance in both quality and diversity. Coverage (COV), on the other hand, quantifies the proportion of ground-truth reference point clouds that correspond to at least one generated shape. Although COV is highly sensitive to mode dropping and serves as an effective measure of diversity, it does not assess the geometric quality of the generated point clouds.

\paragraph{Implementation.} To ensure consistency, we adopt the implementation from DiT-3D and TopoDiT-3D~\citep{mo2023dit3d,Guan25}. The input voxel size is set to $32 \times 32 \times 32 \times 3 (V=32)$, and we train the models for $10,000$ epochs using the Adam optimizer with a learning rate of $1\times 10^{-4}$ and a batch size of $512$.  In our experiments, we set $T = 1000$ and use $S/4$ configuration with a patch size $p = 4$. Specifically, we pick the scale set $\mathcal{S} = \{3,4,5\}$ and momentum set $\mathcal{Q} = \{1,2,3\}$. We pick the window length $w=64$ to generate the HMC token and set the depth of the perceiver resampler to 4 with bottleneck latent dimension $K_b = 128$. After training the DiT backbone, we report metrics from two versions of the proposed algorithm: in the first version, we construct a bank of HMC tokens from the training data, which is then randomly sampled from during generation; in the second version, we separately train a variational autoencoder on the training data and then use the VAE to generate HMC tokens during inference. The VAE baseline uses latent dimension 64,
encoder hidden widths $(2048,1024,512)$ with SiLU activations,
and a decoder with the reversed hidden widths. Its objective
uses $\beta_{\mathrm{VAE}}=0.05$ and unit reconstruction weights.
VAE training uses AdamW with initial learning rate $10^{-3}$,
zero weight decay, batch size 256, cosine learning-rate decay,
gradient clipping at 1, and 2000 epochs.

\subsection{Comparison to Prior Works}
We compare HilbertGen-3D with representative point-cloud generative
models in Table~\ref{tab:shape_metrics}, reporting two variants:
Ours-VAE uses VAE-generated HMC conditions, while Ours-bank uses
bank-sampled HMC conditions. Both variants improve 1-NNA relative
to TopoDiT-3D~\citep{Guan25} across all three categories under both
CD and EMD. Here, improvements in 1-NNA refer to reductions in
the absolute distance from 50, rather than decreases in the raw
accuracy. Coverage gains are more dependent on the category and
conditioning source, with neither variant uniformly outperforming
the other.

For Chair, Ours-VAE achieves the best 1-NNA among all compared
methods, reaching 50.25 under CD and 48.42 under EMD. These
values reduce the absolute distance from 50 by 0.88 and 1.89
percentage points, respectively, compared with TopoDiT-3D.
Ours-bank provides higher coverage than Ours-VAE under both
distance measures and achieves the highest COV-CD and COV-EMD of 57.62 and 57.98,
exceeding TopoDiT-3D by 0.95 and 0.86 percentage points, respectively.

For Airplane, both variants outperform TopoDiT-3D on all four
metrics. Ours-bank achieves the best 1-NNA-CD and 1-NNA-EMD
of 54.79 and 52.83, reducing the corresponding distances from
50 by 2.36 and 3.00 percentage points relative to TopoDiT-3D.
Ours-VAE instead achieves the highest coverage, with COV-CD
and COV-EMD of 58.17 and 59.25, respectively. These represent
improvements of 1.12 and 5.14 percentage points over
TopoDiT-3D. The two condition sources therefore favor different
evaluation criteria even within the same category.

For Car, Ours-VAE achieves the best 1-NNA-CD and 1-NNA-EMD
of 51.14 and 49.81. Their absolute distances from 50 are only
1.14 and 0.19, corresponding to improvements of 6.81 and 8.90
percentage points over TopoDiT-3D. Ours-bank achieves the
highest coverage under both distance measures, reaching
56.82 for COV-CD and 57.86 for COV-EMD, which exceed
TopoDiT-3D by 1.24 and 5.88 percentage points, respectively.
The VAE variant does not share this coverage advantage:
its COV-CD of 53.12 is below TopoDiT-3D, and its COV-EMD
of 54.45 is below LION~\citep{zeng2022lion}.

Overall, Ours-VAE and Ours-bank each attain the best reported values in six of the twelve columns in \Cref{tab:shape_metrics}. VAE conditioning yields stronger 1-NNA results on Chair and Car, whereas bank conditioning provides higher coverage on these categories; the pattern is reversed on Airplane. We attribute this to the intrinsic geometric properties of the categories. We present qualitative comparisons in \Cref{fig:main-result}, with additional category results in \Cref{appendix:additional_result}. More visualizations of point clouds for all categories can be found in \Cref{appendix:more_visual}. 

\begin{table*}[t]
\centering
\caption{Experimental results on 3D point cloud generation across chair,
airplane, and car categories. Metric values are reported as percentages.
HilbertGen-3D (VAE) uses VAE-generated HMC conditions, while HilbertGen-3D (bank) uses
bank-sampled HMC conditions.
For 1-NNA, values closer to 50 are better; for COV, higher values are better.
The best entries in each column are shown in bold, including ties.
Superscripts for both variants report improvements relative to TopoDiT-3D
in percentage points: reductions in absolute distance from 50 for 1-NNA,
and increases for COV.
Positive improvements are green, negative changes are red,
and ties are gray.}
\label{tab:shape_metrics}

\setlength{\tabcolsep}{2pt}
\renewcommand{\arraystretch}{1.15}

\resizebox{\textwidth}{!}{%
\begin{tabular}{l|cccc|cccc|cccc}
\hline
\multirow{3}{*}{Method}
& \multicolumn{4}{c|}{Chair}
& \multicolumn{4}{c|}{Airplane}
& \multicolumn{4}{c}{Car} \\
\cline{2-13}
& \multicolumn{2}{c}{1-NNA ($50$ is optimal)}
& \multicolumn{2}{c|}{COV (\textcolor{green!60!black}{$\uparrow$})}
& \multicolumn{2}{c}{1-NNA ($50$ is optimal)}
& \multicolumn{2}{c|}{COV (\textcolor{green!60!black}{$\uparrow$})}
& \multicolumn{2}{c}{1-NNA ($50$ is optimal)}
& \multicolumn{2}{c}{COV (\textcolor{green!60!black}{$\uparrow$})} \\
& CD & EMD & CD & EMD
& CD & EMD & CD & EMD
& CD & EMD & CD & EMD \\
\hline

r-GAN~\citep{achlioptas2018learning}
& 83.69 & 99.70 & 24.27 & 15.13
& 98.40 & 96.79 & 30.12 & 14.32
& 94.46 & 99.01 & 19.03 & 6.539 \\

l-GAN (CD)
& 68.58 & 83.84 & 41.99 & 29.31
& 87.30 & 93.95 & 38.52 & 21.23
& 66.49 & 88.78 & 38.92 & 23.58 \\

l-GAN (EMD)
& 71.90 & 64.65 & 38.07 & 44.86
& 89.49 & 76.91 & 38.27 & 38.52
& 71.16 & 66.19 & 37.78 & 45.17 \\

PointFlow~\citep{yang2019pointflow}
& 62.84 & 60.57 & 42.90 & 50.00
& 75.68 & 70.74 & 47.90 & 46.41
& 58.10 & 56.25 & 46.88 & 50.00 \\

SoftFlow~\citep{kim2020softflow}
& 59.21 & 60.05 & 41.39 & 47.43
& 76.05 & 65.80 & 46.91 & 47.90
& 64.77 & 60.09 & 42.90 & 44.60 \\

SetVAE~\citep{kim2021setvae}
& 58.84 & 60.57 & 46.83 & 44.26
& 76.54 & 67.65 & 43.70 & 48.40
& 59.94 & 59.94 & 49.15 & 46.59 \\

DPF-Net~\citep{Klokov20}
& 62.00 & 58.53 & 44.71 & 48.79
& 75.18 & 65.55 & 46.17 & 48.89
& 62.35 & 54.48 & 45.74 & 49.43 \\
\hline

DPM~\citep{luo2021diffusion}
& 60.05 & 74.77 & 44.86 & 35.50
& 76.42 & 86.91 & 48.64 & 33.83
& 68.89 & 79.97 & 44.03 & 34.94 \\

PVD~\citep{Zhou20213DSG}
& 57.09 & 60.87 & 36.68 & 49.24
& 73.82 & 64.81 & 48.88 & 52.09
& 54.55 & 53.83 & 41.19 & 50.56 \\

LION~\citep{zeng2022lion}
& 53.70 & 52.34 & 48.94 & 52.11
& 67.41 & 61.23 & 47.16 & 49.63
& 53.41 & 51.14 & 50.00 & 56.53 \\
\hline

GET3D~\citep{gao2022get3d}
& 75.26 & 72.49 & 43.36 & 42.77
& - & - & - & -
& 75.26 & 72.49 & 15.04 & 18.38 \\

MeshDiffusion~\citep{liu2023meshdiffusion}
& 53.69 & 57.63 & 46.00 & 46.71
& 66.44 & 76.26 & 47.34 & 42.15
& 81.43 & 87.84 & 34.07 & 25.85 \\

DiT-3D~\citep{mo2023dit3d}
& 56.31 & 55.82 & 47.21 & 50.75
& 66.21 & 63.17 & 48.26 & 45.89
& 68.13 & 64.24 & 43.44 & 44.08 \\

TopoDiT-3D~\citep{Guan25}
& 48.87 & 46.53 & 56.67 & 57.12
& 57.15 & 55.83 & 57.05 & 54.11
& 57.95 & 59.09 & 55.58 & 51.98 \\
\hline

HilbertGen-3D (VAE)
% Chair
& \textbf{50.25}\textsuperscript{\textcolor{green!60!black}{$+0.88$}}
& \textbf{48.42}\textsuperscript{\textcolor{green!60!black}{$+1.89$}}
& 54.32\textsuperscript{\textcolor{red}{$-2.35$}}
& 54.83\textsuperscript{\textcolor{red}{$-2.29$}}

% Airplane
& 56.79\textsuperscript{\textcolor{green!60!black}{$+0.36$}}
& 53.25\textsuperscript{\textcolor{green!60!black}{$+2.58$}}
& \textbf{58.17}\textsuperscript{\textcolor{green!60!black}{$+1.12$}}
& \textbf{59.25}\textsuperscript{\textcolor{green!60!black}{$+5.14$}}
% Car
& \textbf{51.14}\textsuperscript{\textcolor{green!60!black}{$+6.81$}}
& \textbf{49.81}\textsuperscript{\textcolor{green!60!black}{$+8.90$}}
& 53.12\textsuperscript{\textcolor{red}{$-2.46$}}
& 54.45\textsuperscript{\textcolor{green!60!black}{$+2.47$}} \\
HilbertGen-3D (bank)
% Chair
& 49.01\textsuperscript{\textcolor{green!60!black}{$+0.14$}}
& 47.36\textsuperscript{\textcolor{green!60!black}{$+0.83$}}
& \textbf{57.62}\textsuperscript{\textcolor{green!60!black}{$+0.95$}}
& \textbf{57.98}\textsuperscript{\textcolor{green!60!black}{$+0.86$}}

% Airplane
& \textbf{54.79}\textsuperscript{\textcolor{green!60!black}{$+2.36$}}
& \textbf{52.83}\textsuperscript{\textcolor{green!60!black}{$+3.00$}}
& 57.92\textsuperscript{\textcolor{green!60!black}{$+0.87$}}
& 58.58\textsuperscript{\textcolor{green!60!black}{$+4.47$}}
% Car
& 46.21\textsuperscript{\textcolor{green!60!black}{$+4.16$}}
& 47.54\textsuperscript{\textcolor{green!60!black}{$+6.63$}}
& \textbf{56.82}\textsuperscript{\textcolor{green!60!black}{$+1.24$}}
& \textbf{57.86}\textsuperscript{\textcolor{green!60!black}{$+5.88$}} \\

\hline
\end{tabular}%
}
\end{table*}

\begin{table}[t]
\centering
\caption{Ablation study on the \textit{Airplane} category. For 1-NNA, values closer to 50 is better, and for COV, higher values are better.}
\label{tab:chair_ablation}

\setlength{\tabcolsep}{6pt}
\renewcommand{\arraystretch}{1.15}

\begin{tabular}{l|cc|cc}
\hline
\multirow{2}{*}{Method}
& \multicolumn{2}{c|}{1-NNA($50$ is optimal)}
& \multicolumn{2}{c}{COV(\textcolor{green!60!black}{$\uparrow$})} \\
& CD & EMD & CD & EMD \\
\hline

Ours
&  \textbf{55.57 }& 53.66 & \textbf{60.98} & \textbf{59.27} \\

w/o HMC tokens
& 57.33 & 54.76 & 58.50 & 57.67 \\

w/o global HMC embedding
& 55.84 & \textbf{53.58} & 60.24 & 58.25 \\

w/o bottleneck fusion
& 60.00 & 54.54 & 55.92 & 55.83 \\

\hline
\end{tabular}
\end{table}

\subsection{Ablation analysis.}
We conduct ablation studies on the Airplane category to examine
the contributions of Hilbert serialization, generalized-dimension
descriptors, and bottleneck fusion. Similar to the sensitivity study, we use a batch size of 128 and train for 2000 epochs.
Table~\ref{tab:chair_ablation} compares the full model with
variants that remove each component. The full model achieves
the best 1-NNA-CD and both coverage metrics, while the variant
without generalized dimensions achieves the closest
1-NNA-EMD to 50. Throughout this comparison, changes in
1-NNA are interpreted through the absolute distance from 50,
rather than the raw accuracy alone.

Removing the HMC tokens worsens all four reported metrics. The 1-NNA-CD and 1-NNA-EMD increase from 55.57 and 53.66 to 57.33 and 54.76, respectively, increasing their distances from 50 by 1.76 and 1.10 percentage points. Meanwhile, COV-CD decreases from 60.98 to 58.50, and COV-EMD decreases from 59.27 to 57.67, corresponding to reductions of 2.48 and 1.60 percentage points. These results support the use of Hilbert serialization for constructing spatially structured conditioning signals. Among the tested ablations, removing this component produces the largest deterioration in EMD-based 1-NNA.

Removing the global HMC embedding increases 1-NNA-CD from 55.57 to 55.84, moving it 0.27 percentage points farther from 50. COV-CD and COV-EMD decrease from 60.98 and 59.27 to 60.24 and 58.25, respectively, corresponding to reductions of 0.74 and 1.02 percentage points. However, 1-NNA-EMD improves slightly from 53.66 to 53.58, reducing its distance from 50 by 0.08 percentage points. Thus, the global descriptor provides complementary benefits for CD-based distribution matching and coverage in this experiment, but does not improve every reported metric.

Removing bottleneck fusion produces the largest deterioration in CD-based 1-NNA and both coverage metrics. The 1-NNA-CD increases from 55.57 to 60.00, moving 4.43 percentage points farther from 50, while COV-CD and COV-EMD decrease to 55.92 and 55.83, representing reductions of 5.06 and 3.44 percentage points. The 1-NNA-EMD also worsens, increasing from 53.66 to 54.54 and moving 0.88 percentage points farther from 50. These results indicate that the fusion architecture is important for using HMC conditions effectively in the evaluated configuration.

Overall, all three components contribute to the full model's
CD-based 1-NNA and coverage advantages over the ablated
variants. Bottleneck fusion has the largest observed effect
on these three metrics, while the EMD-based 1-NNA results
show a more nuanced pattern. The ablations therefore
support the combined design without establishing that
every component improves every evaluation criterion.

\section{Conclusion}
In this work, we presented HilbertGen-3D, a diffusion framework for point cloud generation based on Hilbert--Multifractal Conditioning (HMC). Experiments suggest that HMC is a promising conditioning strategy for improving global coherence and structural fidelity in generated 3D shapes. More broadly, the proposed framework shows that locality-preserving spatial serializations together with multi-resolution scaling statistics can serve as an effective alternative to purely local conditioning schemes. On the other hand, it shall be noted that, while HilbertGen-3D can effectively generate point clouds in settings when prior information of the overall density distribution can be exploited, alternative approaches (such as persistent homology) may perform well otherwise. %However, we note that HMC is not a replacement for persistent homology when exact topological invariance is required. 
Future directions include distilling the diffusion model to reduce computational cost~\citep{yin2026recursiveselfimprovementonpolicydistillation} and using Bayesian techniques to further improve generation accuracy~\citep {yueh2026bayesianlearningepisodiczerosum}. 

\clearpage

\section{Acknowledgement}
Ziyi Zhang was supported by Lee-Stanziale Ohana Fellowship. Guannan Qu was supported by NSF Grants 2339112, 2512805, AFOSR Grant FA9550261B281, Jane Street, Pennsylvania Infrastructure
Technology Alliance, CMU Manufacturing Futures Institute. Yorie Nakahira was supported by National Science Foundation (NSF) Grant No. ECCS-2442948 (CAREER) and Japan Science and Technology Agency (JST): Grant No. JPMJPR2516 (PRESTO).
Paul Bogdan was supported by the National Science Foundation (NSF) under NSF Award 2243104 under the Center for Complex Particle Systems (COMPASS), CAREER Award CPS-1453860, NSF Mid-Career Advancement Award BCS-2527046, U.S. Army Research Office (ARO) under Grant No. W911NF-23-1-0111, Defense Advanced Research Projects Agency (DARPA) Young Faculty Award and DARPA Director Fellowship Award under Grant Number N66001-17-1-4044, DARPA grant HR0011263E153, Intel Faculty Awards, and a Northrop Grumman grant. P.B. is also grateful to the National Institutes of Health (NIH) for Grants R01 AG079957 and RF1 AG082201. The views, opinions, and/or findings in this article are those of the authors and should not be interpreted as official views or policies of the Department of Defense, the National Institutes of Health, or the National Science Foundation.

\bibliographystyle{plainnat}
\bibliography{ref}

@book{mandelbrot2013multifractals,
  title={Multifractals and 1/ƒ noise: Wild self-affinity in physics (1963--1976)},
  author={Mandelbrot, Benoit B},
  year={2013},
  publisher={Springer}
}

@InProceedings{Hao_2021_ICCV,
    author    = {Hao, Zekun and Mallya, Arun and Belongie, Serge and Liu, Ming-Yu},
    title     = {GANcraft: Unsupervised 3D Neural Rendering of Minecraft Worlds},
    booktitle = {Proceedings of the IEEE/CVF International Conference on Computer Vision (ICCV)},
    month     = {October},
    year      = {2021},
    pages     = {14072-14082}
}

@article{Litany2017DeformableSC,
  title={Deformable Shape Completion with Graph Convolutional Autoencoders},
  author={Or Litany and Alexander M. Bronstein and Michael M. Bronstein and Ameesh Makadia},
  journal={2018 IEEE/CVF Conference on Computer Vision and Pattern Recognition},
  year={2017},
  pages={1886-1895}
}

@article{Gao2023MaskedDT,
  title={Masked Diffusion Transformer is a Strong Image Synthesizer},
  author={Shanghua Gao and Pan Zhou and Mingg-Ming Cheng and Shuicheng Yan},
  journal={2023 IEEE/CVF International Conference on Computer Vision (ICCV)},
  year={2023},
  pages={23107-23116}
}

@article{Guan25,
author = {Guan, Zechao and Yan, Feng and Du, Shuai and Ma, Lin and Liu, Qingshan},
year = {2025},
month = {05},
pages = {},
title = {TopoDiT-3D: Topology-Aware Diffusion Transformer with Bottleneck Structure for 3D Point Cloud Generation},
doi = {10.48550/arXiv.2505.09140}
}

@article{Salat17,
title = {Multifractal methodology},
author = {Hadrien Salat and Roberto Murcio and Elsa Arcaute},
journal = {Physica A: Statistical Mechanics and its Applications},
volume = {473},
pages = {467-487},
year = {2017},
issn = {0378-4371},
doi = {https://doi.org/10.1016/j.physa.2017.01.041},
url = {https://www.sciencedirect.com/science/article/pii/S0378437117300341}
}

@article{xu2023survey,
  author  = {Xu, Qun-Ce and Mu, Tai-Jiang and Yang, Yong-Liang},
  title   = {A Survey of Deep Learning-Based 3D Shape Generation},
  journal = {Computational Visual Media},
  volume  = {9},
  pages   = {407--442},
  year    = {2023},
  doi     = {10.1007/s41095-022-0321-5}
}

@inproceedings{yang2019pointflow,
  author    = {Yang, Guandao and Huang, Xun and Hao, Zekun and Liu, Ming-Yu and Belongie, Serge and Hariharan, Bharath},
  title     = {PointFlow: 3D Point Cloud Generation With Continuous Normalizing Flows},
  booktitle = {Proceedings of the IEEE/CVF International Conference on Computer Vision (ICCV)},
  pages     = {4541--4550},
  year      = {2019}
}

@inproceedings{cai2020shapegf,
  author    = {Cai, Ruojin and Yang, Guandao and Averbuch-Elor, Hadar and Hao, Zekun and Belongie, Serge and Snavely, Noah and Hariharan, Bharath},
  title     = {Learning Gradient Fields for Shape Generation},
  booktitle = {Computer Vision -- ECCV 2020},
  pages     = {364--381},
  year      = {2020}
}

@inproceedings{luo2021diffusion,
  author    = {Luo, Shitong and Hu, Wei},
  title     = {Diffusion Probabilistic Models for 3D Point Cloud Generation},
  booktitle = {Proceedings of the IEEE/CVF Conference on Computer Vision and Pattern Recognition (CVPR)},
  pages     = {2837--2845},
  year      = {2021}
}

@inproceedings{zeng2022lion,
  author    = {Zeng, Xiaohui and Vahdat, Arash and Williams, Francis and Gojcic, Zan and Litany, Or and Fidler, Sanja and Kreis, Karsten},
  title     = {LION: Latent Point Diffusion Models for 3D Shape Generation},
  booktitle = {Advances in Neural Information Processing Systems},
  year      = {2022}
}

@inproceedings{gao2022get3d,
  author    = {Gao, Jun and Shen, Tianchang and Wang, Zian and Chen, Wenzheng and Yin, Kangxue and Li, Daiqing and Litany, Or and Gojcic, Zan and Fidler, Sanja},
  title     = {GET3D: A Generative Model of High Quality 3D Textured Shapes Learned from Images},
  booktitle = {Advances in Neural Information Processing Systems},
  year      = {2022}
}

@article{nichol2022pointe,
  author  = {Nichol, Alex and Jun, Heewoo and Dhariwal, Prafulla and Mishkin, Pamela and Chen, Mark},
  title   = {Point-E: A System for Generating 3D Point Clouds from Complex Prompts},
  journal = {arXiv preprint arXiv:2212.08751},
  year    = {2022}
}

@article{jun2023shape,
  author  = {Jun, Heewoo and Nichol, Alex},
  title   = {Shap-E: Generating Conditional 3D Implicit Functions},
  journal = {arXiv preprint arXiv:2305.02463},
  year    = {2023}
}

@inproceedings{mo2023dit3d,
  author    = {Mo, Shentong and Xie, Enze and Chu, Ruihang and Hong, Lanqing and Niessner, Matthias and Li, Zhenguo},
  title     = {DiT-3D: Exploring Plain Diffusion Transformers for 3D Shape Generation},
  booktitle = {Advances in Neural Information Processing Systems},
  year      = {2023}
}

@article{moon2001hilbert,
  author  = {Moon, Bongki and Jagadish, H. V. and Faloutsos, Christos and Saltz, Joel H.},
  title   = {Analysis of the Clustering Properties of the Hilbert Space-Filling Curve},
  journal = {IEEE Transactions on Knowledge and Data Engineering},
  volume  = {13},
  number  = {1},
  pages   = {124--141},
  year    = {2001}
}

@article{wang2023octformer,
  author  = {Wang, Peng-Shuai},
  title   = {OctFormer: Octree-based Transformers for 3D Point Clouds},
  journal = {ACM Transactions on Graphics},
  volume  = {42},
  number  = {4},
  pages   = {1--11},
  year    = {2023},
  doi     = {10.1145/3592131}
}

@inproceedings{wu2024ptv3,
  author    = {Wu, Xiaoyang and Jiang, Li and Wang, Peng-Shuai and Liu, Zhijian and Liu, Xihui and Qiao, Yu and Ouyang, Wanli and He, Tong and Zhao, Hengshuang},
  title     = {Point Transformer V3: Simpler Faster Stronger},
  booktitle = {Proceedings of the IEEE/CVF Conference on Computer Vision and Pattern Recognition (CVPR)},
  pages     = {4840--4851},
  year      = {2024}
}

@inproceedings{liang2024pointmamba,
  author    = {Liang, Dingkang and Zhou, Xin and Xu, Wei and Zhu, Xingkui and Zou, Zhikang and Ye, Xiaoqing and Tan, Xiao and Bai, Xiang},
  title     = {PointMamba: A Simple State Space Model for Point Cloud Analysis},
  booktitle = {Advances in Neural Information Processing Systems},
  year      = {2024}
}

@article{pandhare2023point2point,
  author  = {Pandhare, Athrva Atul},
  title   = {Point2Point: A Framework for Efficient Deep Learning on Hilbert Sorted Point Clouds with Applications in Spatio-Temporal Occupancy Prediction},
  journal = {arXiv preprint arXiv:2306.16306},
  year    = {2023}
}

@misc{
achlioptas2018learning,
title={Learning Representations and Generative Models for 3D Point Clouds},
author={Panos Achlioptas and Olga Diamanti and Ioannis Mitliagkas and Leonidas Guibas},
year={2018},
url={https://openreview.net/forum?id=BJInEZsTb},
}

@article{Zhou20213DSG,
  title={3D Shape Generation and Completion through Point-Voxel Diffusion},
  author={Linqi Zhou and Yilun Du and Jiajun Wu},
  journal={2021 IEEE/CVF International Conference on Computer Vision (ICCV)},
  year={2021},
  pages={5806-5815}
}

@inproceedings{kim2020softflow,
  title={Softflow: Probabilistic framework for normalizing flow on manifolds},
  author={Kim, Hyeongju and Lee, Hyeonseung and Kang, Woo Hyun and Lee, Joun Yeop and Kim, Nam Soo},
  booktitle={Advances in Neural Information Processing Systems},
  volume={33},
  pages={17865--17876},
  year={2020}
}

@inproceedings{kim2021setvae,
  title={Setvae: Learning hierarchical composition for generative modeling of set-structured data},
  author={Kim, Jinwoo and Yoo, Jaehoon and Lee, Juho and Hong, Seunghoon},
  booktitle={Proceedings of the IEEE/CVF Conference on Computer Vision and Pattern Recognition},
  pages={15168--15177},
  year={2021}
}

@inproceedings{Klokov20,
author = {Klokov, Roman and Boyer, Edmond and Verbeek, Jakob},
title = {Discrete Point Flow Networks for Efficient Point Cloud Generation},
year = {2020},
isbn = {978-3-030-58591-4},
publisher = {Springer-Verlag},
address = {Berlin, Heidelberg},
doi = {10.1007/978-3-030-58592-1_41},
booktitle = {ECCV 2020},
pages = {694–710},
numpages = {17}
}

@inproceedings{
liu2023meshdiffusion,
title={MeshDiffusion: Score-based Generative 3D Mesh Modeling},
author={Zhen Liu and Yao Feng and Michael J. Black and Derek Nowrouzezahrai and Liam Paull and Weiyang Liu},
booktitle={The Eleventh International Conference on Learning Representations },
year={2023}
}

@techreport{shapenet2015,
  title       = {{ShapeNet: An Information-Rich 3D Model Repository}},
  author      = {Chang, Angel X. and Funkhouser, Thomas and Guibas, Leonidas and Hanrahan, Pat and Huang, Qixing and Li, Zimo and Savarese, Silvio and Savva, Manolis and Song, Shuran and Su, Hao and Xiao, Jianxiong and Yi, Li and Yu, Fisher},
  number      = {arXiv:1512.03012 [cs.GR]},
  institution = {Stanford University --- Princeton University --- Toyota Technological Institute at Chicago},
  year        = {2015}
}

@inproceedings{Ho20denoise,
 author = {Ho, Jonathan and Jain, Ajay and Abbeel, Pieter},
 booktitle = {Advances in Neural Information Processing Systems},
 editor = {H. Larochelle and M. Ranzato and R. Hadsell and M.F. Balcan and H. Lin},
 pages = {6840--6851},
 publisher = {Curran Associates, Inc.},
 title = {Denoising Diffusion Probabilistic Models},
 url = {https://proceedings.neurips.cc/paper_files/paper/2020/file/4c5bcfec8584af0d967f1ab10179ca4b-Paper.pdf},
 volume = {33},
 year = {2020}
}

@misc{yin2026recursiveselfimprovementonpolicydistillation,
      title={Recursive Self-Improvement via On-Policy Distillation for Reasoning}, 
      author={Shangjian Yin and Zehao Zhao and Kavosh Asadi and Rui Liu and Yuchen Lu and Shike Mei and Hang Cui and Luke Simon and Zhouxing Shi and Hamed Firooz},
      year={2026},
      eprint={2609.30652},
      archivePrefix={arXiv},
      primaryClass={cs.CL},
      url={https://arxiv.org/abs/2609.30652}, 
}

@misc{yueh2026bayesianlearningepisodiczerosum,
      title={Bayesian Learning in Episodic Zero-Sum Games}, 
      author={Chang-Wei Yueh and Andy Zhao and Ashutosh Nayyar and Rahul Jain},
      year={2026},
      eprint={2603.20604},
      archivePrefix={arXiv},
      primaryClass={cs.LG},
      url={https://arxiv.org/abs/2603.20604}, 
}

@misc{zhang2026learningfractionalorderdynamicssingle,
      title={Learning Fractional-Order Dynamics from a Single Trajectory}, 
      author={Xiaole Zhang and Ziyi Zhang and Zehao Zhao and Stephen Tu and Guannan Qu and Yorie Nakahira and Paul Bogdan},
      year={2026},
      eprint={2609.18127},
      archivePrefix={arXiv},
      primaryClass={cs.LG},
      url={https://arxiv.org/abs/2609.18127}, 
}

@misc{zhang2025samplingcomplexityawareframeworkdiscretetime,
      title={A Sampling Complexity-aware Framework for Discrete-time Fractional-Order Dynamical System Identification}, 
      author={Xiaole Zhang and Vijay Gupta and Paul Bogdan},
      year={2025},
      eprint={2501.17499},
      archivePrefix={arXiv},
      primaryClass={eess.SY},
      url={https://arxiv.org/abs/2501.17499}, 
}

%%%%%%%%%%%%%%%%%%%%%%%%%%%%%%%%%%%%%%%%%%%%%%%%%%%%%%%%%%%%
\clearpage
\appendix

\section{Additional results}
\label{appendix:additional_result}

\begin{table*}[ht]
\centering
\caption{Results on the additional Lamp and Vessel categories.
All metric values are reported as percentages.
Ours-VAE uses VAE-generated HMC conditions, while Ours-Bank
uses bank-sampled HMC conditions.
For 1-NNA, values closer to 50 are better; for COV, higher
values are better.
The best entries in each column are shown in bold, including ties.
Green superscripts report improvements relative to TopoDiT-3D
in percentage points: reductions in absolute distance from 50
for 1-NNA, and increases for COV.}
\label{tab:additional_appendix}
\scriptsize
\setlength{\tabcolsep}{2pt}
\renewcommand{\arraystretch}{1.15}

\begin{tabular}{l|cccc|cccc}
\hline
\multirow{3}{*}{Method}
& \multicolumn{4}{c|}{Lamp}
& \multicolumn{4}{c}{Vessel} \\
\cline{2-9}
& \multicolumn{2}{c}{1-NNA ($50$ optimal)}
& \multicolumn{2}{c|}{COV (\textcolor{green!60!black}{$\uparrow$})}
& \multicolumn{2}{c}{1-NNA ($50$ optimal)}
& \multicolumn{2}{c}{COV (\textcolor{green!60!black}{$\uparrow$})} \\
& CD & EMD & CD & EMD
& CD & EMD & CD & EMD \\
\hline

TopoDiT-3D
& 58.83 & 57.75 & 50.43 & 49.13
& 72.21 & 78.96 & 32.81 & 28.12 \\

Ours-VAE
& \textbf{49.28}\textsuperscript{\textcolor{green!60!black}{+8.11}}
& \textbf{48.78}\textsuperscript{\textcolor{green!60!black}{+6.53}}
& \textbf{55.75}\textsuperscript{\textcolor{green!60!black}{+5.32}}
& 56.61\textsuperscript{\textcolor{green!60!black}{+7.48}}
& \textbf{56.08}\textsuperscript{\textcolor{green!60!black}{+16.13}}
& 51.74\textsuperscript{\textcolor{green!60!black}{+27.22}}
& 49.31\textsuperscript{\textcolor{green!60!black}{+16.50}}
& 53.12\textsuperscript{\textcolor{green!60!black}{+25.00}} \\

Ours-Bank
& 47.49\textsuperscript{\textcolor{green!60!black}{+6.32}}
& \textbf{48.78}\textsuperscript{\textcolor{green!60!black}{+6.53}}
& 55.60\textsuperscript{\textcolor{green!60!black}{+5.17}}
& \textbf{58.19}\textsuperscript{\textcolor{green!60!black}{+9.06}}
& 56.51\textsuperscript{\textcolor{green!60!black}{+15.70}}
& \textbf{49.05}\textsuperscript{\textcolor{green!60!black}{+28.01}}
& \textbf{54.17}\textsuperscript{\textcolor{green!60!black}{+21.36}}
& \textbf{58.33}\textsuperscript{\textcolor{green!60!black}{+30.21}} \\

\hline
\end{tabular}
\end{table*}

We further evaluate HilbertGen-3D on the Lamp and Vessel categories
in Table~\ref{tab:additional_appendix}, reporting both VAE-generated
and bank-sampled HMC conditions. Both variants outperform
TopoDiT-3D~\citep{Guan25} on all four metrics in both categories.
As in the main comparison, improvements in 1-NNA are measured
by reductions in the absolute distance from 50, rather than
decreases in the raw accuracy.

For Lamp, Ours-VAE achieves the best 1-NNA-CD of 49.28,
reducing the distance from 50 by 8.11 percentage points relative
to TopoDiT-3D. Both variants attain a 1-NNA-EMD of 48.78,
corresponding to an improvement of 6.53 percentage points.
Ours-VAE also obtains the highest COV-CD of 55.75, compared
with 55.60 for Ours-Bank and 50.43 for TopoDiT-3D.
Ours-Bank instead achieves the highest COV-EMD of 58.19,
exceeding TopoDiT-3D by 9.06 percentage points and Ours-VAE
by 1.58 percentage points. Thus, the VAE variant provides
better CD-based 1-NNA and slightly higher CD-based coverage,
while the bank variant provides higher EMD-based coverage.

For Vessel, Ours-VAE achieves the best 1-NNA-CD of 56.08,
reducing the distance from 50 by 16.13 percentage points
relative to TopoDiT-3D. Ours-Bank achieves the best
1-NNA-EMD of 49.05, whose distance from 50 is only 0.95,
compared with 1.74 for Ours-VAE. These correspond to
improvements of 28.01 and 27.22 percentage points,
respectively, over TopoDiT-3D. Both variants also substantially
increase coverage. Ours-Bank achieves the highest COV-CD
and COV-EMD of 54.17 and 58.33, improving upon
TopoDiT-3D by 21.36 and 30.21 percentage points.
Ours-VAE reaches 49.31 and 53.12, respectively, which
remain above the baseline but below the bank-conditioned
results.

\begin{figure}[hbt!]
    \centering
    \includegraphics[width=\linewidth]{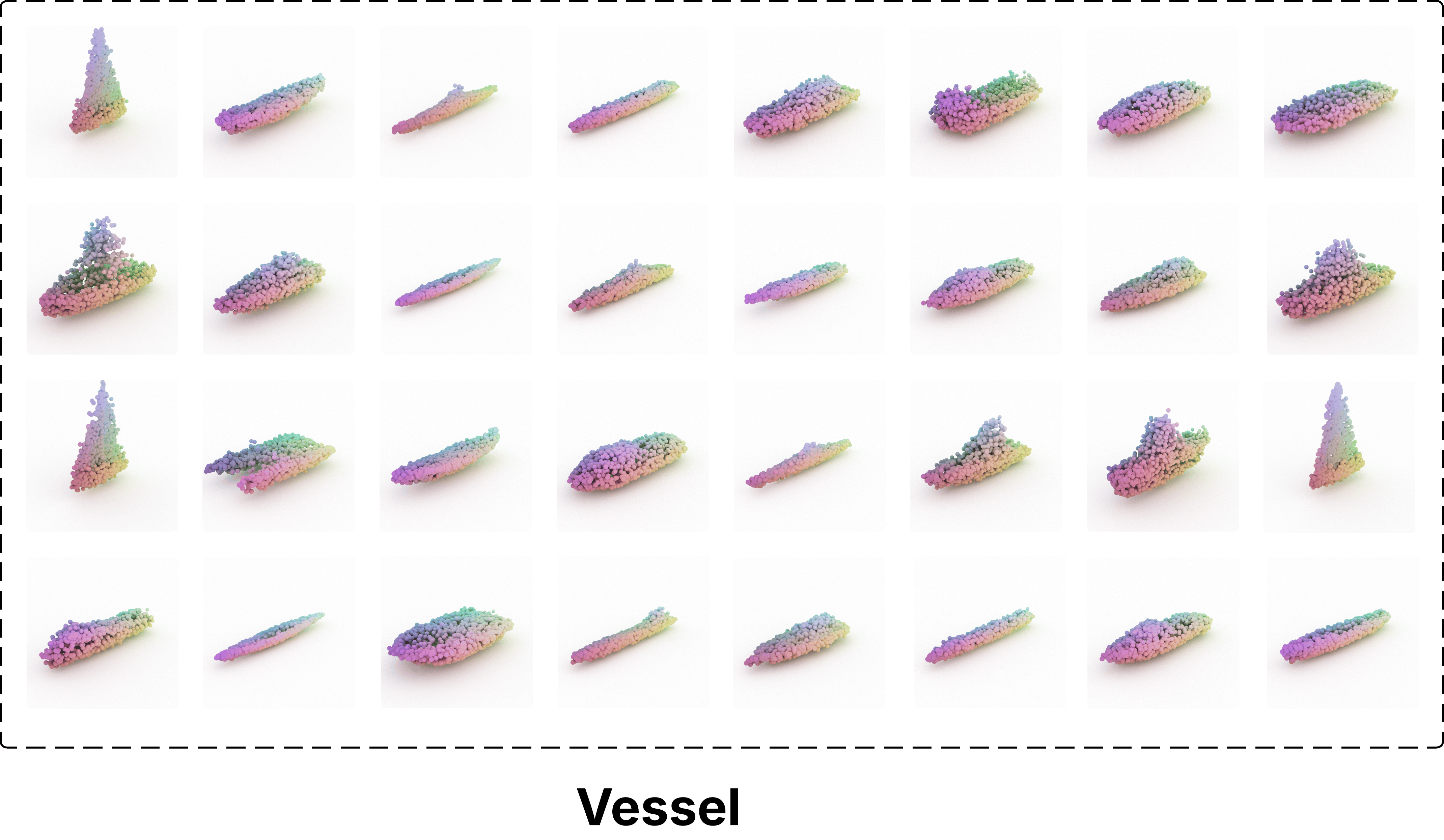}
    \caption{HilbertGen-3d samples on vessel.}
    \label{fig:result_vessel}
\end{figure}

\begin{figure}[hbt!]
    \centering
    \includegraphics[width=\linewidth]{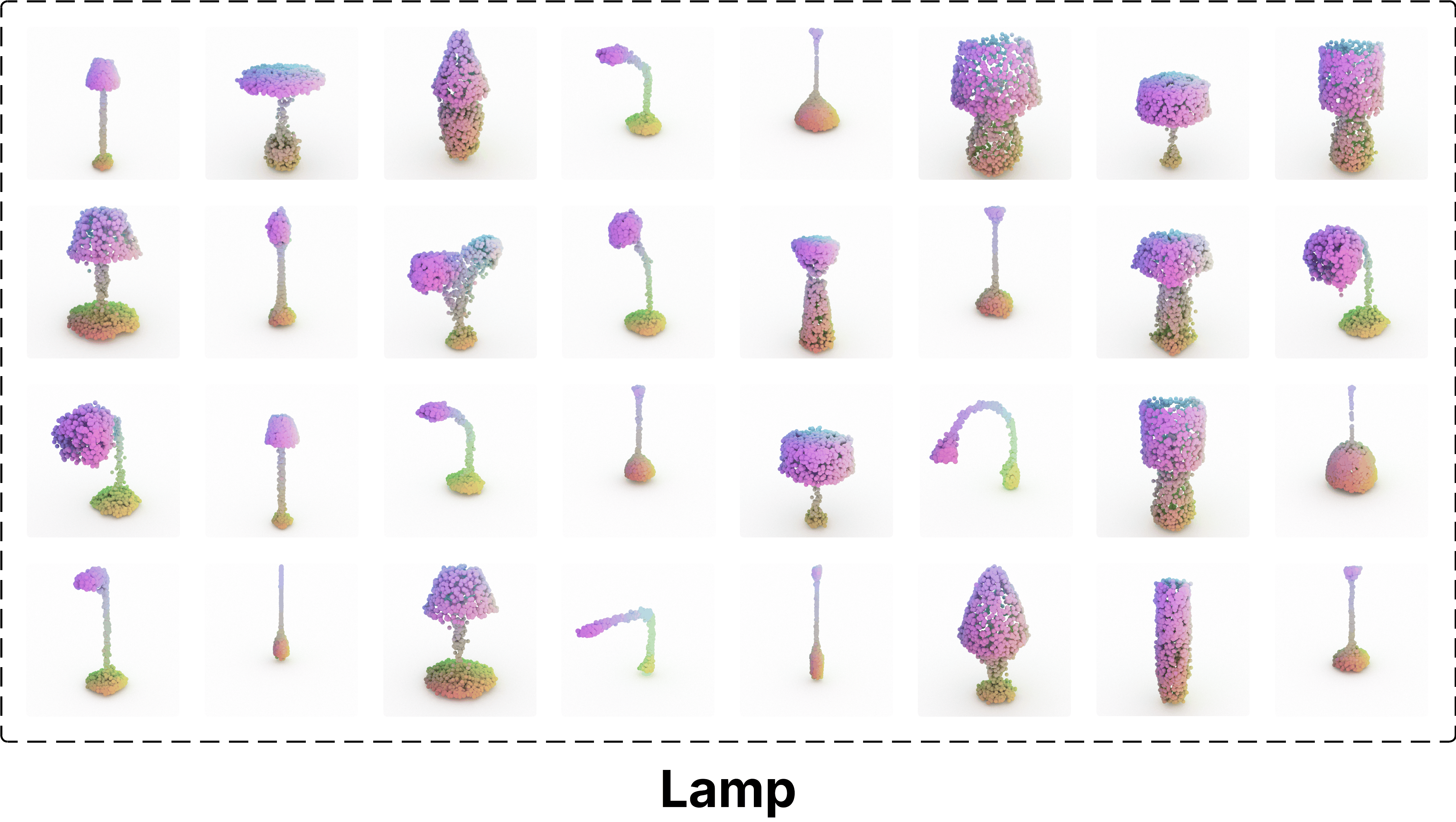}
    \caption{HilbertGen-3d samples on lamp.}
    \label{fig:result_lamp}
\end{figure}

Overall, these additional-category experiments support the
effectiveness of HMC conditioning with either a learned
condition prior or a training-set condition bank. The VAE
variant achieves better CD-based 1-NNA on both categories,
while the bank variant achieves higher EMD-based coverage
on both and higher CD-based coverage on Vessel.
Neither conditioning source is uniformly superior across
the reported metrics. The full visualizations of these experiments appear in \Cref{fig:result_lamp} and \Cref{fig:result_vessel}.

\section{Full Parameter Sensitivity Results}
\label{app:full_sensitivity}

\begin{table}[hbt!]
\centering
\caption{Parameter sensitivity analysis on the \textit{Airplane} category (mean over four evaluation seeds: 0, 11, 12, 13). All runs use the normalized training recipe with a batch size of 128 for 2000 epochs; conditions are sampled from the train-split HMC bank (Ours-Bank) and scored under the validation protocol with $N=664$ samples, exact EMD, and no size correction. The baseline setting of each group is marked with $\dagger$. 1-NNA is better when closer to 50, while higher COV is better. The best entry in each column within a group is shown in bold, including ties. All values are multiplied by 100 for readability.}
\label{tab:sensitivity_airplane_mean}
\small
\setlength{\tabcolsep}{6pt}
\renewcommand{\arraystretch}{1.1}
\begin{tabular}{llcccc}
\toprule
Parameter & Setting & \multicolumn{2}{c}{1-NNA ($50$ optimal)} & \multicolumn{2}{c}{COV($\uparrow$)} \\
\cmidrule(lr){3-4} \cmidrule(lr){5-6}
& & CD & EMD & CD & EMD \\
\midrule
Resampler depth & 1 & 63.67 & 55.75 & 55.35 & 56.88 \\
 & 2 & 66.03 & 64.99 & 48.58 & 46.10 \\
 & 3 & 59.45 & \textbf{53.09} & 56.77 & 57.98 \\
 & 4$^\dagger$ & \textbf{55.57} & 53.66 & \textbf{60.98} & \textbf{59.27} \\
\midrule
MLP ratio & 2.5 & 55.96 & 53.01 & 61.02 & 59.02 \\
 & 3 & \textbf{55.46} & \textbf{52.27} & 61.17 & \textbf{59.58} \\
 & 3.5 & 56.32 & 53.53 & \textbf{61.19} & 59.35 \\
 & 4$^\dagger$ & 55.57 & 53.66 & 60.98 & 59.27 \\
\midrule
Bottleneck latents & 32 & 56.72 & 55.40 & \textbf{60.98} & 58.81 \\
 & 64 & 56.57 & 52.43 & 59.19 & 58.65 \\
 & 96 & 55.58 & \textbf{51.95} & 60.75 & \textbf{59.48} \\
 & 128$^\dagger$ & \textbf{55.57} & 53.66 & \textbf{60.98} & 59.27 \\
\midrule
Scale set & \{3,4\} & 56.15 & 52.52 & 60.52 & 59.10 \\
 & \{3,5\} & 56.04 & \textbf{51.91} & 60.33 & 58.79 \\
 & \{3,4,5\}$^\dagger$ & \textbf{55.57} & 53.66 & \textbf{60.98} & \textbf{59.27} \\
\midrule
$q$ orders & \{1\} & 56.08 & 53.11 & \textbf{61.31} & 59.17 \\
 & \{1,2\} & 57.80 & 55.59 & 61.17 & 57.77 \\
 & \{1,2,3\}$^\dagger$ & \textbf{55.57} & 53.66 & 60.98 & \textbf{59.27} \\
 & \{1,2,3,4\} & 56.99 & \textbf{52.65} & 59.31 & 58.29 \\
\bottomrule
\end{tabular}
\end{table}

We provide the complete sensitivity results in Table~\ref{tab:sensitivity_airplane_mean}. 
All experiments are conducted on the \textit{Airplane} category with a batch size of 128 and training for 2000 epochs. 

\paragraph{Resampler depth.}
Resampler depth has the largest observed effect among the
parameter ranges considered. Depth 4 achieves the best
1-NNA-CD of 55.57 and the highest COV-CD and COV-EMD
of 60.98 and 59.27 within this group.
Compared with depth 3, it reduces the distance of 1-NNA-CD
from 50 by 3.88 percentage points and increases COV-CD
and COV-EMD by 4.21 and 1.29 percentage points.
However, depth 3 achieves the closest 1-NNA-EMD to 50,
at 53.09, compared with 53.66 for depth 4.
The weaker results at depth 2 further show that performance
does not improve monotonically with depth.
These observations support depth 4 for the coverage and
CD-based criteria in this experiment, while retaining an
EMD-based 1-NNA advantage for depth 3.

\paragraph{MLP ratio.}
An MLP ratio of 3 achieves the best 1-NNA-CD,
1-NNA-EMD, and COV-EMD within its group, at 55.46,
52.27, and 59.58, respectively.
Relative to the ratio-4 baseline, the distances of the two
1-NNA metrics from 50 decrease by 0.11 and 1.39
percentage points, while COV-CD and COV-EMD increase
by 0.19 and 0.31 percentage points.
Ratio 3.5 yields the highest COV-CD of 61.19, exceeding
ratio 3 by only 0.02 percentage points.
Thus, the largest tested MLP ratio is not favored by the
reported means: ratio 3 improves every metric relative
to the baseline and is close to the group maximum in
the remaining coverage comparison. We have chosen MLP ratio of 4 in the hope that smaller step-size, larger batch size, and larger epochs would benefit a slightly more complicated model. 

\paragraph{Number of bottleneck latents.}
Increasing the bottleneck length does not yield uniform gains.
The baseline with 128 latents attains the best 1-NNA-CD
of 55.57, but the result with 96 latents is nearly identical
at 55.58. In contrast, 96 latents achieves the best
1-NNA-EMD and COV-EMD, at 51.95 and 59.48.
Compared with 128 latents, this reduces the EMD-based
1-NNA distance from 50 by 1.71 percentage points and
increases COV-EMD by 0.21 percentage points, while
COV-CD decreases by 0.23 percentage points.
The 32- and 128-latent settings tie for the highest
COV-CD at the reported precision, although the
32-latent setting has worse 1-NNA under both distances.
The results therefore show a trade-off between bottleneck
length and the different evaluation criteria, rather than
a consistent advantage for the largest bottleneck.

\paragraph{Scale selection.}
The full scale set $\mathcal S=\{3,4,5\}$ achieves the
best 1-NNA-CD and both coverage metrics within its group.
However, the two-scale alternatives yield better
1-NNA-EMD, with $\{3,5\}$ attaining the closest value
to 50 at 51.91. Compared with $\{3,5\}$, the full
scale set improves the CD-based 1-NNA distance by
0.47 percentage points and increases COV-CD and
COV-EMD by 0.65 and 0.48 percentage points,
but worsens the EMD-based 1-NNA distance by 1.75
percentage points.
Accordingly, retaining all three scales favors coverage
and CD-based distribution matching in this experiment,
but does not improve every measure of distribution
matching. Since every tested configuration contains
multiple scales, this comparison does not establish
an advantage over single-scale conditioning.

\paragraph{Moment-order selection.}
The baseline moment set $\mathcal Q=\{1,2,3\}$
achieves the best 1-NNA-CD and COV-EMD within its
group, at 55.57 and 59.27.
Using only $\mathcal Q=\{1\}$ gives the highest
COV-CD of 61.31 and also yields a 1-NNA-EMD value
closer to 50 than the baseline, at 53.11.
The set $\{1,2,3,4\}$ achieves the best 1-NNA-EMD
of 52.65, improving its distance from 50 by 1.01
percentage points relative to the baseline, but reducing
COV-CD and COV-EMD by 1.67 and 0.98 percentage
points. Adding moment orders therefore does not
produce a monotonic improvement.
The $\{1\}$ configuration retains the information-dimension
descriptor; it should not be interpreted as removing
the global HMC descriptor altogether.

\paragraph{Interpretation and scope.}
The baseline provides a common reference, but the table
does not identify it as uniformly optimal. In particular,
MLP ratio 3 improves all four reported means relative
to that reference. The groupwise comparisons also do
not establish the performance of a configuration that
combines the individually favorable settings, since
interactions between parameter choices are not evaluated
by these comparisons.

These results concern Ours-Bank on Airplane under the
2000-epoch training recipe. They do not establish the
same parameter preferences for Ours-VAE, other
categories, or longer training schedules.
The four seeds are evaluation seeds; the table does
not quantify variation across independently trained
models. Furthermore, without seedwise dispersion
estimates, small differences such as 0.01 in 1-NNA-CD
between 96 and 128 latents, or 0.02 in COV-CD between
MLP ratios 3 and 3.5, should be interpreted as numerical
rankings rather than evidence of statistically significant
performance differences.

\section{HMC-DiT3D for multi-category generation.}

\begin{table}[t]
\centering
\caption{Exploration studies on multi-class training. All models are trained on the listed classes and tested on the chair category. HilbertGen-3D (VAE) samples chair HMC conditions from the chair condition VAE, and HilbertGen-3D (Bank) draws them from the chair training bank. For 1-NNA, values closer to 50 are better; for COV, higher values are better. DiT-3D-S/4 and TopoDiT-3D-S/4 results are quoted from TopoDiT-3D.}
\label{tab:multiclass}
\small
\setlength{\tabcolsep}{4pt}
\renewcommand{\arraystretch}{1.1}
\begin{tabular}{lcccccc}
\toprule
\multirow{2}{*}{Model} & Train & Test & \multicolumn{2}{c}{1-NNA(50 optimal)} & \multicolumn{2}{c}{COV($\uparrow$)} \\
\cmidrule(lr){4-5} \cmidrule(lr){6-7}
 & Class & Class & CD & EMD & CD & EMD \\
\midrule
\multirow{3}{*}{DiT-3D-S/4}
 & Chair & Chair & 51.99 & 49.94 & 54.76 & 57.37 \\
 & Chair,Car & Chair & 69.87 & 65.71 & 45.28 & 52.20 \\
 & Chair,Car,Airplane & Chair & 75.80 & 74.72 & 41.46 & 47.89 \\
\midrule
\multirow{3}{*}{TopoDiT-3D-S/4}
 & Chair & Chair & 49.84 & 46.46 & 58.28 & 60.69 \\
 & Chair,Car & Chair & 60.46 & 60.31 & 56.17 & 56.92 \\
 & Chair,Car,Airplane & Chair & 68.75 & 61.21 & 50.90 & 54.96 \\
\midrule
\multirow{3}{*}{HilbertGen-3D (VAE)-S/4}
 & Chair & Chair & 50.25 & 48.42 & 54.32 & 54.83 \\
 & Chair,Car & Chair & 50.03 & 49.64 & 55.59 & 54.42 \\
 & Chair,Car,Airplane & Chair & 51.93	&48.35	&54.22&	53.96 \\
\midrule
\multirow{3}{*}{HilbertGen-3D (Bank)-S/4}
 & Chair & Chair & 49.01 & 47.36 & 57.62 & 57.98 \\
 & Chair,Car & Chair & 48.81 & 47.79 & 57.93 & 58.59 \\
 & Chair,Car,Airplane & Chair & 49.36	&47.64	&57.72&58.03 \\
\bottomrule
\end{tabular}
\end{table}

To investigate how our model behaves when a single network has to serve several categories, we train HMC-DiT3D-S/4 on \{Chair\}, \{Chair, Car\}, and \{Chair, Car, Airplane\}, respectively, and test every model on the chair category, with the results shown in Table~\ref{tab:multiclass}. For both baselines, adding categories to the training set clearly hurts chair generation: DiT-3D-S/4 drops from 54.76 to 41.46 in COV-CD and its 1-NNA-CD rises from 51.99 to 75.80, while TopoDiT-3D-S/4 drops from 58.28 to 50.90 in COV-CD and its 1-NNA-CD rises from 49.84 to 68.75, since the class label is the only signal that tells these models which category to produce. In contrast, HMC-DiT3D is comparatively stable in multi-class training, particularly with bank-sampled conditions. With bank-sampled conditions, 1-NNA-CD stays within 48.81--49.36 and COV-CD within 57.62--57.93 across the three training sets without a clear upward or downward trend. For VAE-sampled conditions, we see a slight downward trend: 1-NNA-CD increases from 50.25 to 51.93, while COV-EMD decreases from 54.83 to 53.96. However, both methods are still significantly more robust than both benchmarks. We attribute this robustness to the HMC conditioning: the multi-scale Hilbert measure sequence of a chair describes its geometry explicitly, so the denoiser does not need to infer the category from the label alone, and the additional categories do not dilute the chair-specific guidance.

\section{Computational Cost Analysis}
\label{app:computational_cost}

We further evaluate the computational cost of our method and compare it with representative diffusion-transformer baselines, including DiT-3D and TopoDiT-3D. 
Since all our experiments are conducted under the \textbf{S/4} configuration, we focus on computational efficiency under this setting rather than performing an additional model-scale sweep. 
Following prior work, we report the sampling time in seconds per iteration (s/it), and additionally measure the peak GPU memory consumption during training. 
As shown in Table~\ref{tab:computational_cost}, our method achieves a sampling speed of \textbf{2.60 s/it}, which is shorter than the reported S/4 inference time of both DiT-3D and TopoDiT-3D. 
Meanwhile, our method also uses less training memory than both baselines. 
These results indicate that the proposed design does not introduce heavy computational overhead; instead, it maintains efficient inference while reducing the practical memory burden during training.

\begin{table}[t]
\centering
\caption{Computational cost comparison under the S/4 configuration. Sampling time is reported in seconds per iteration. Training memory denotes the peak GPU memory usage during training.}
\label{tab:computational_cost}
\small
\setlength{\tabcolsep}{6pt}
\renewcommand{\arraystretch}{1.12}
\begin{tabular}{lccc}
\toprule
Method & Model Size & Sampling Time (s/it) $\downarrow$ & Training Memory (GB) $\downarrow$ \\
\midrule
DiT-3D      & S/4 & 8.50 & 78.4 \\
TopoDiT-3D  & S/4 & 8.20 & 51.2 \\
Ours        & S/4 & \textbf{2.60} & \textbf{51.0} \\
\bottomrule
\end{tabular}
\end{table}

We also compare the training throughput on the chair category, which is the main category used in our controlled analysis. 
As shown in Table~\ref{tab:training_epoch_time}, our method requires \textbf{2.3 seconds per epoch}, while the baseline setting requires \textbf{11.8 seconds per epoch}. 
This corresponds to a \textbf{5.13$\times$ training speedup}, or equivalently a \textbf{80.5\% reduction} in per-epoch training time. 
Together with the lower peak memory usage in Table~\ref{tab:computational_cost}, this shows that our method improves computational efficiency not only during sampling, but also during training.

\begin{table}[t]
\centering
\caption{Training time comparison on the chair category under the S/4 configuration. Epoch time is reported in seconds per epoch.}
\label{tab:training_epoch_time}
\small
\setlength{\tabcolsep}{8pt}
\renewcommand{\arraystretch}{1.12}
\begin{tabular}{lccc}
\toprule
Method & Model Size & Epoch Time (s/epoch) $\downarrow$ & Relative Speedup \\
\midrule
Topodit & S/4 & 11.8 & 1.00$\times$ \\
Ours     & S/4 & \textbf{2.3} & \textbf{5.13$\times$} \\
\bottomrule
\end{tabular}
\end{table}

\section{Result Visualization}
\label{appendix:more_visual}

\begin{figure}[hbt!]
    \centering
    \includegraphics[width=\linewidth]{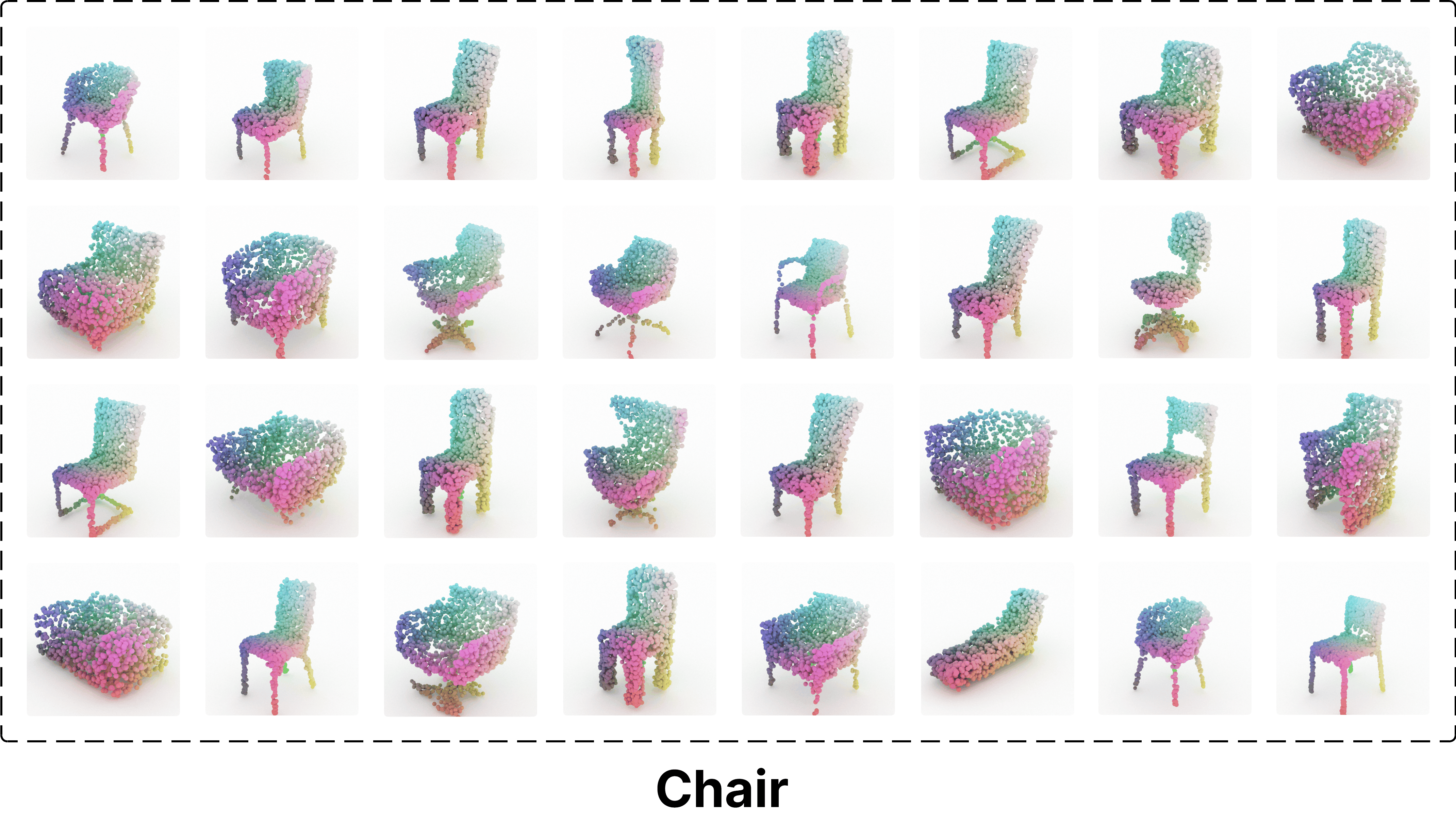}
    \caption{HilbertGen-3d samples on Chair.}
    \label{fig:result_chair}
\end{figure}

\begin{figure}[hbt!]
    \centering
    \includegraphics[width=\linewidth]{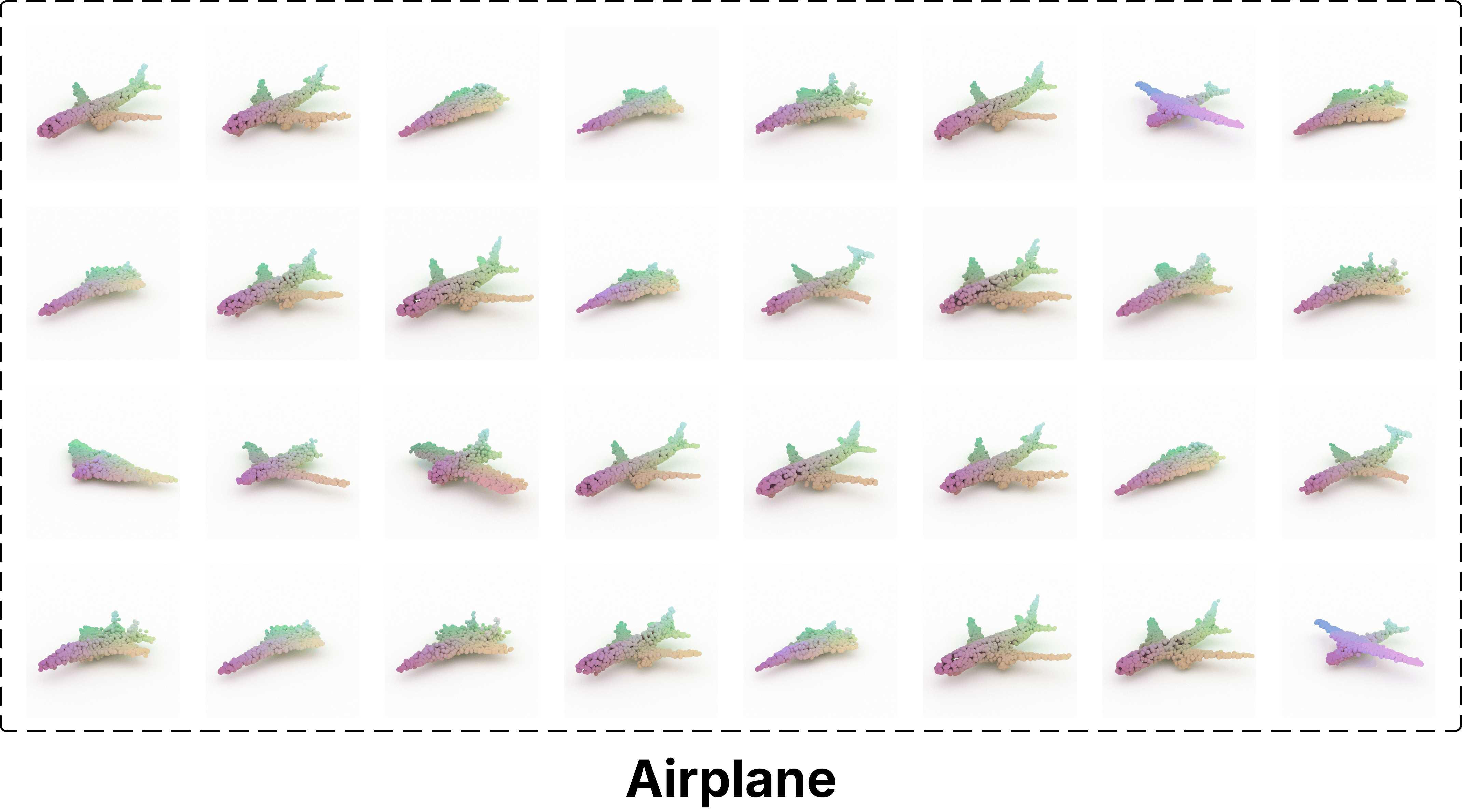}
    \caption{HilbertGen-3d samples on airplane.}
    \label{fig:result_airplane}
\end{figure}
\begin{figure}[hbt!]
    \centering
    \includegraphics[width=\linewidth]{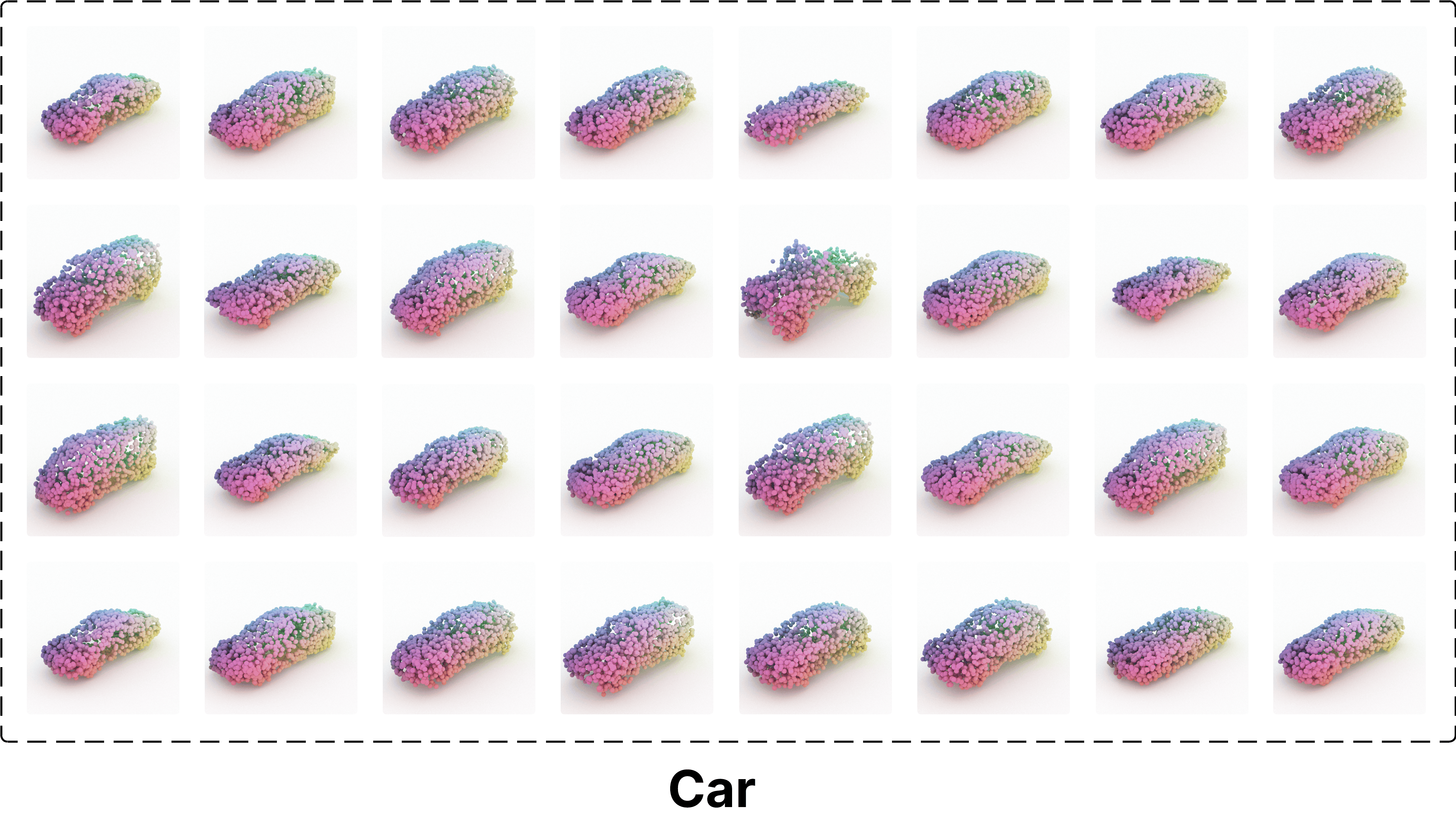}
    \caption{HilbertGen-3d samples on car.}
    \label{fig:result_car}
\end{figure}

\clearpage
%%%%%%%%%%%%%%%%%%%%%%%%%%%%%%%%%%%%%%%%%%%%%%%%%%%%%%%%%%%%
\section{Qualitative visualizations of denoising process}

\begin{figure}[hbt!]
    \centering
    \includegraphics[width=0.90\linewidth]{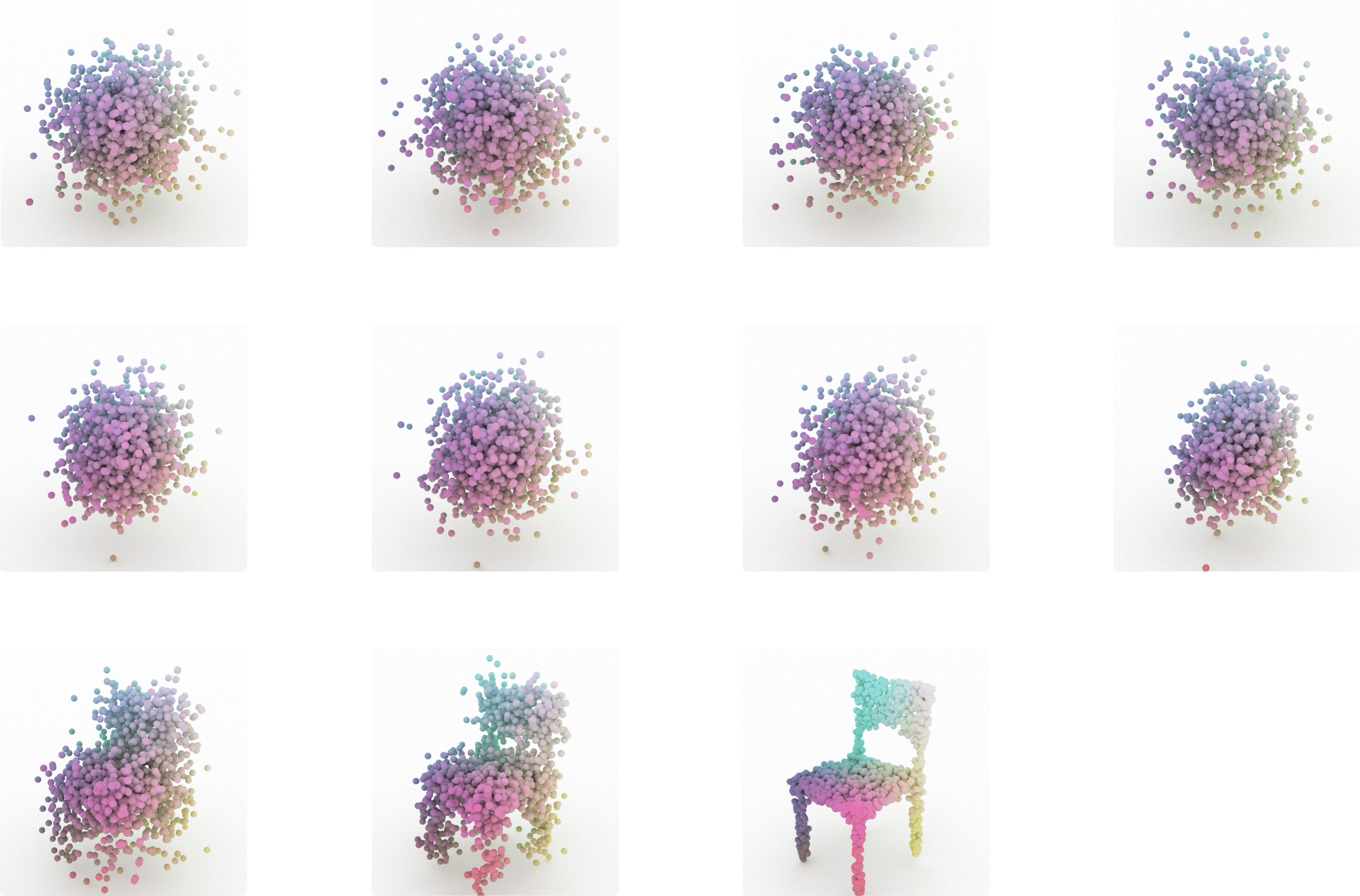}
    \caption{Qualitative visualizations of denoising process for chair generation.}
    \label{fig:dif_chair}
\end{figure}

\begin{figure}[hbt!]
    \centering
    \includegraphics[width=0.90\linewidth]{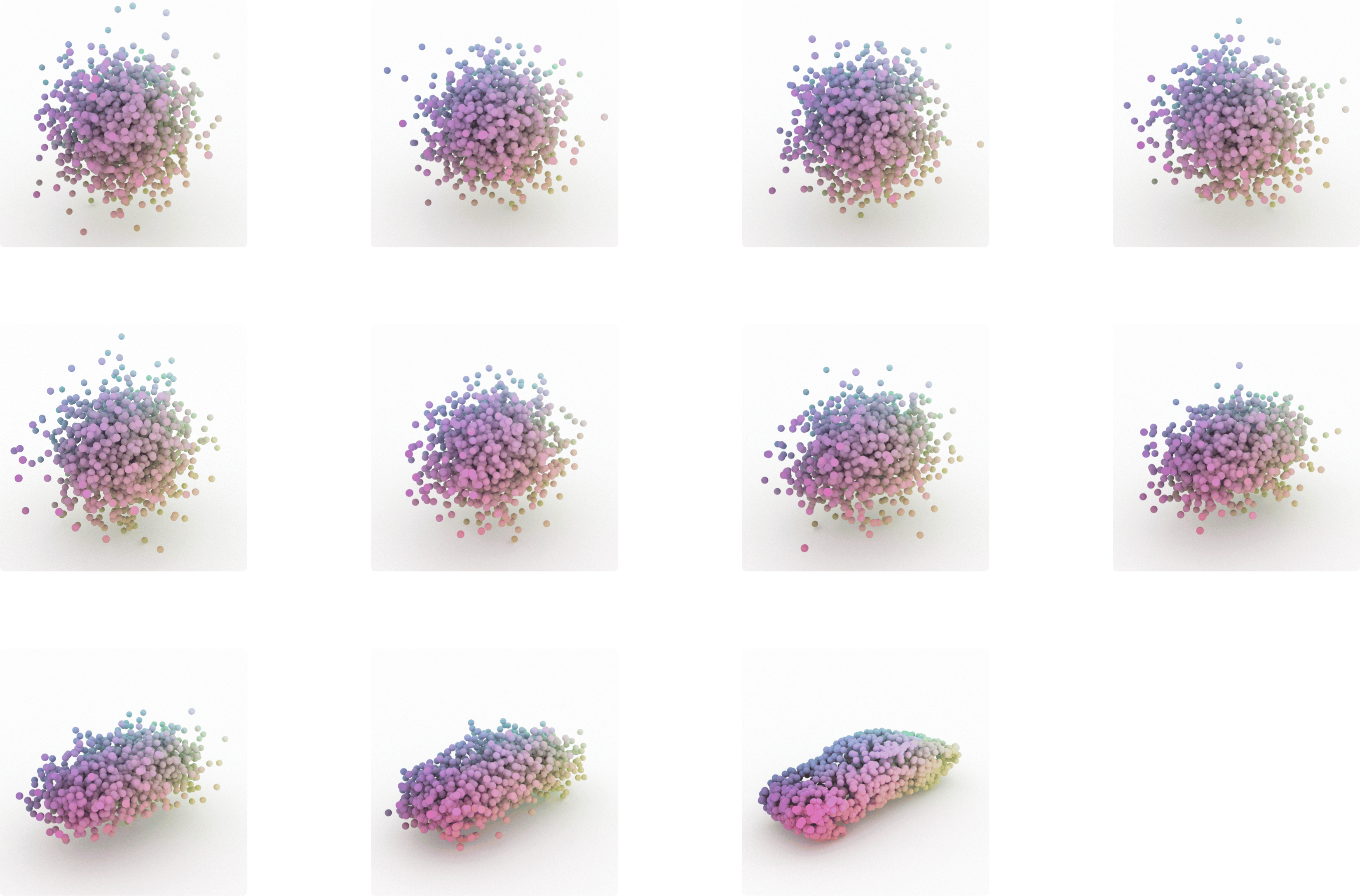}
    \caption{Qualitative visualizations of denoising process for car generation.}
    \label{fig:dif_car}
\end{figure}

\begin{figure}[hbt!]
    \centering
    \includegraphics[width=0.90\linewidth]{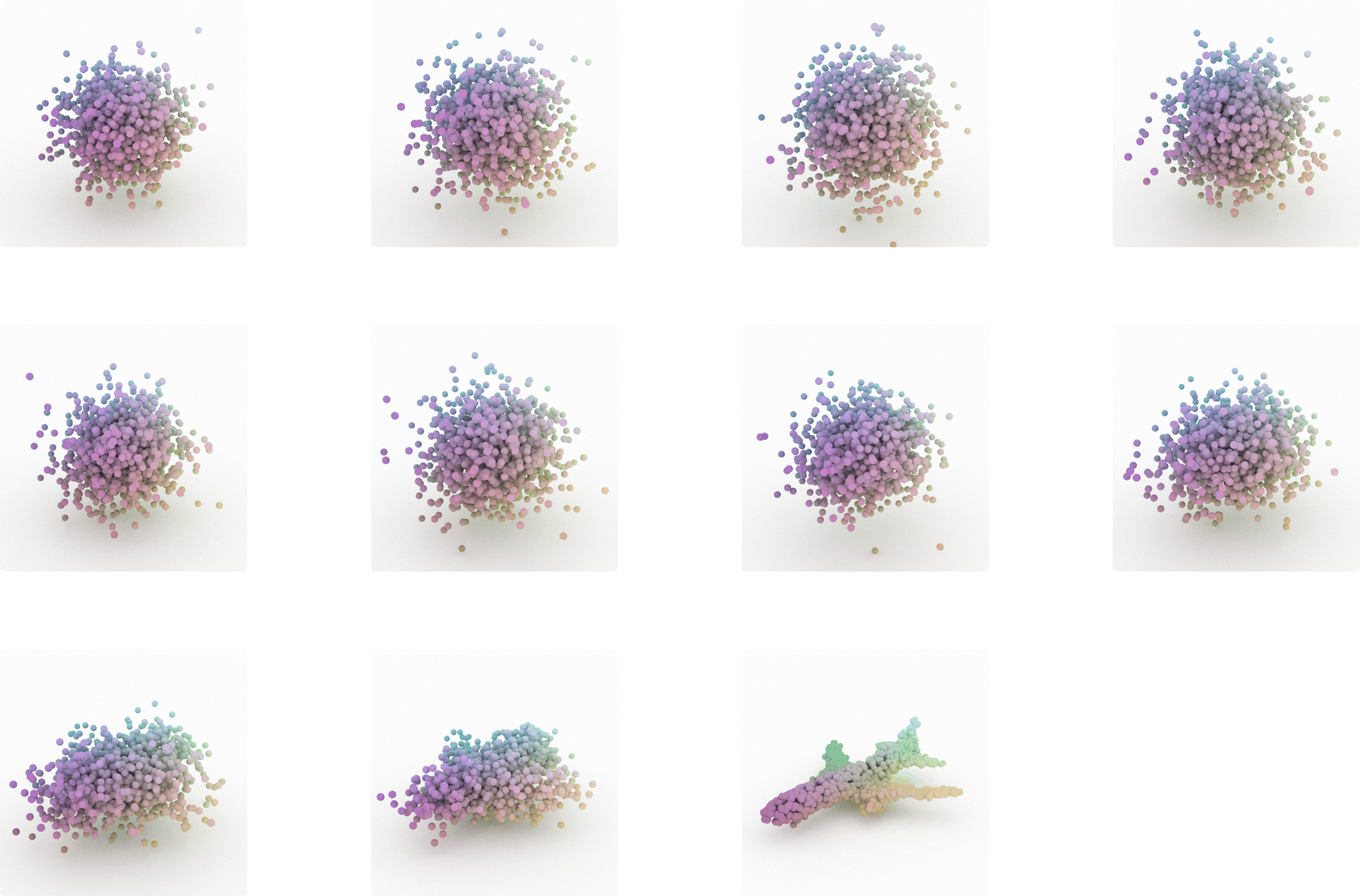}
    \caption{Qualitative visualizations of denoising process for airplane generation.}
    \label{fig:dif_plane}
\end{figure}

%\clearpage
%\input{checklist.tex}

\end{document}